%% file: combined.tex
\documentclass[10pt,twocolumn,letterpaper]{article}

\usepackage{iccv}
\usepackage{times}
\usepackage{epsfig}
\usepackage{graphicx}
\usepackage{amsmath}
\usepackage{amssymb}
\usepackage{authblk}
\usepackage{placeins}
\usepackage[symbol*]{footmisc}
\usepackage{multicol}
\usepackage{fancyhdr}
\usepackage{float} 

\usepackage{graphicx}
\usepackage{booktabs}
\usepackage{multicol}
\usepackage{multirow}
\usepackage{enumitem}
\usepackage{etoolbox, siunitx}

\usepackage{amssymb}
\usepackage{pifont}
\usepackage[pagebackref=true,breaklinks=true,letterpaper=true,colorlinks,bookmarks=false]{hyperref}

\usepackage{fancyhdr}

\usepackage[capitalize]{cleveref}
\crefname{section}{Sec.}{Secs.}
\Crefname{section}{Section}{Sections}
\Crefname{table}{Table}{Tables}
\crefname{table}{Tab.}{Tabs.}

\iccvfinalcopy 

\def\iccvPaperID{11050} 

\ificcvfinal\fi

\begin{document}

\title{TEGLO: High Fidelity Canonical Texture Mapping from Single-View Images}

\author{Vishal Vinod$^{1,2,}$\thanks{These authors contributed equally to this work.} \;\; Tanmay Shah$^{2,*}$ \; Dmitry Lagun$^2$\\
$^1$University of California, San Diego \; $^2$Google Research\\
{\tt\small vvinod@ucsd.edu, \{shaht, dlagun\}@google.com}\\
{\tt\small \href{https://teglo-nerf.github.io/}{https://teglo-nerf.github.io/}}}


\maketitle
\ificcvfinal\thispagestyle{empty}\fi

\begin{abstract}
    
    
    Recent work in Neural Fields (NFs) learn 3D representations from class-specific single view image collections. However, they are unable to reconstruct the input data preserving high-frequency details. Further, these methods do not disentangle appearance from geometry and hence are not suitable for tasks such as texture transfer and editing. In this work, we propose TEGLO (Textured EG3D-GLO) for learning 3D representations from single view in-the-wild image collections for a given class of objects. We accomplish this by training a conditional Neural Radiance Field (NeRF) without any explicit 3D supervision. We equip our method with editing capabilities by creating a dense correspondence mapping to a 2D canonical space. We demonstrate that such mapping enables texture transfer and texture editing without requiring meshes with shared topology. Our key insight is that by mapping the input image pixels onto the texture space we can achieve near perfect reconstruction ($\geq 74$ dB PSNR at $1024^2$ resolution). Our formulation allows for high quality 3D consistent novel view synthesis with high-frequency details at megapixel image resolution. 
    
    
    \vspace{-5mm}
\end{abstract}
\section{Introduction}
\vspace{-1mm}
\label{sec:intro}
    \input{Sections/1_Intro.tex}

\section{Related Work}
\vspace{-1mm}
\label{sec:related}
    \input{Sections/2_Related.tex}

\vspace{-1mm}
\section{Proposed Method}
\label{sec:method}
    \input{Sections/3_Methods.tex}

\vspace{-1.5mm}
\section{Experiments and Results}
\vspace{-1mm}
\label{sec:experiments}
    \input{Sections/4_Experiments.tex}

\vspace{-1mm}
\section{Discussion}
\label{sec:conclusion}
\vspace{-0.5mm}
    \input{Sections/5_Discussion.tex}
    
\vspace{-1mm}
\section{Conclusion}
\label{sec:conclusion}
    \input{Sections/6_Conclusion.tex}

{\small
\bibliographystyle{ieee_fullname}
\bibliography{egbib}
}

\clearpage

\setcounter{equation}{0}
\setcounter{figure}{0}
\setcounter{table}{0}
\setcounter{section}{0}
\renewcommand{\thefigure}{S\arabic{figure}}
\renewcommand{\thesection}{S\arabic{section}}  
\renewcommand{\thetable}{S\arabic{table}}
\renewcommand{\theequation}{S\arabic{equation}}

\renewcommand{\theHfigure}{S\arabic{figure}}
\renewcommand{\theHsection}{S\arabic{section}}  
\renewcommand{\theHtable}{S\arabic{table}}
\renewcommand{\theHequation}{S\arabic{equation}}



\onecolumn
\vskip .375in
\begin{center}
{\Large \bf {\normalsize \textit{Supplementary material for}} \\ TEGLO: High Fidelity Canonical Texture Mapping from Single-View Images \par}
\vspace*{32pt}
\end{center}

\begin{multicols}{2} 



\section{Network architecture details}
\label{sec:network}
TEGLO Stage-1 denoted as $\mathcal{N}$ consists of a latent table, StyleGANv2 \cite{karras2019style} generator layers, a 2-layer tri-plane decoder, a differentiable volume rendering module and a light weight camera predictor. For the experiments in this work, we use 512-dimensional latents for the latent table. Following \cite{chan2022efficient}, the output from the StyleGANv2 generator is of shape $256\times256\times96$ giving us $\mathrm{k}=32$-channel tri-planes. Note that despite the use of a fixed size tri-plane, TEGLO enables arbitrary resolution synthesis as it employs a GLO-based auto-decoder training strategy. As noted in the main paper, this also enables single-view 3D reconstruction at any resolution. Table.(\ref{tab:camera-pred}), includes the details for the camera predictor network.

\vspace{-2mm}

\begin{table}[H]
    \centering
    \caption{Camera Predictor.}
    \label{tab:camera-pred}
    \begin{tabular}{@{}lcccc@{}}
    \hline
         Layer & Kernel & Filters & Stride & Activation\\
    \hline
    Conv 2D & 3 & 128 & 2 & LeakyReLU\\
    Conv 2D & 3 & 64 & 2 & LeakyReLU\\
    Conv 2D & 3 & 32 & 2 & LeakyReLU\\ 
    Conv 2D & 3 & 16 & 2 & LeakyReLU\\ 
    Dense & - & 25 & - & Linear\\ 
    \hline
    \end{tabular}
    \vspace{-2mm}
\end{table}

TEGLO Stage-2 consists of a latent mapping network ($\mathcal{L}$), a dense correspondence network ($\mathcal{M}$) and a basis network ($\mathcal{C}$). We describe the network details in Fig.(\ref{fig:teglo-stage-2-network}). $\mathcal{M}$ is a LipMLP \cite{liu2022learning} with a Lipschitz regularizer at every layer to encourage Lipschitz continuity with respect to the inputs. Note that the same network implementation is used for all experiments - FFHQ \cite{karras2019style}, CelebA-HQ \cite{liu2015faceattributes, karras2017progressive}, AFHQv2-Cats \cite{karras2021alias, choi2020stargan} and SRN-Cars \cite{shapenet2015, chang2015shapenet}. 

\vspace{-1mm}
\section{Training details}
\label{sec:suppl-impl}
In TEGLO Stage-1, to train $\mathcal{N}$, we use the single-view image dataset and the approximate pose obtained from the shape-matching least-squares optimization described in Sec.(\ref{sec:experiments}) in the main paper following the procedure in \cite{rebain2022lolnerf}. We train $\mathcal{N}$ for 500K steps using the Adam optimizer \cite{kingma2014adam} on 8 NVIDIA V100 GPUs (16 GB VRAM each) taking a total of 46 hours to complete training at $256^2$ resolution. We also employ an exponential learning rate decay from $5e$-4 to $1e$-4. In each train step, we use latent $w_i$ to condition the NeRF to reconstruct the object ($o_i \in \mathcal{I}$). Then, we use the loss $\mathcal{L}_{\mathcal{N}} = \mathcal{L}_{\text{RGB}} + \mathcal{L}_{\text{Perceptual}} + \mathcal{L}_{\text{Camera}}$ to train the network.

\begin{figure}[H]
  \centering
  \scalebox{0.9}{
  \includegraphics[width=0.85\linewidth]{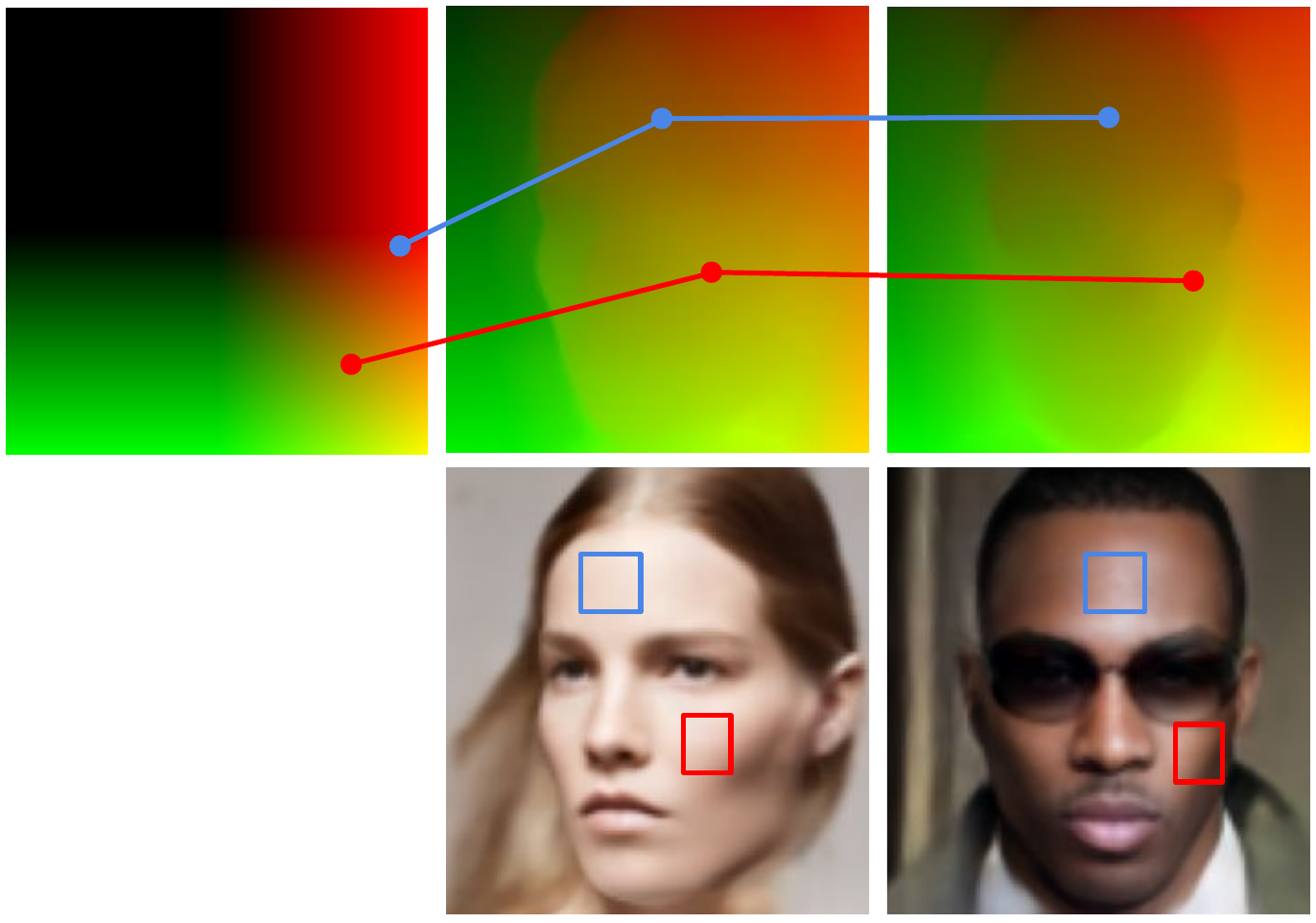}
  }
  \caption{\textbf{Correspondence maps} - Establishing dense correspondence between 3D objects in the 2D  canonical co-ordinate space.}
  \label{fig:suppl-corr}
\end{figure}

\begin{figure*}[t]
  \centering
  \includegraphics[width=0.99\linewidth]{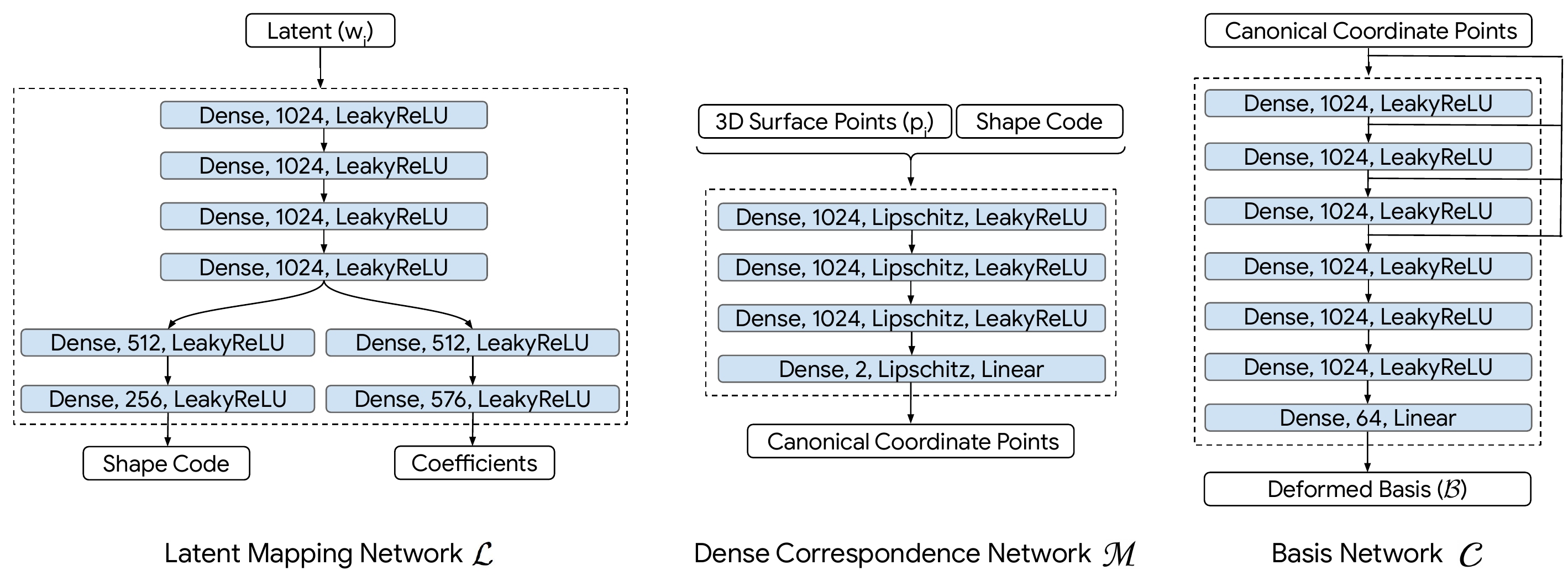}
  \caption{\textbf{TEGLO Stage-2 network details}}
  \label{fig:teglo-stage-2-network}
  \vspace{-2mm}
\end{figure*}

To train TEGLO Stage-2, we first render a dataset $\mathcal{D}$ with five camera views: $e_i = \{ v_f, v_l, v_r, v_t, v_b \}$ for that object. In this work, we render 1000 identities for $\mathcal{D}$ giving us 5000 total views. We train TEGLO Stage-2 for 1 Million steps (1000 epochs) using the Adam \cite{kingma2014adam} optimizer on 8 NVIDIA V100 GPUs (16 GB VRAM each) taking a total of 50 hours to complete training. In each train step, we use the latent $w_i$ as input to $\mathcal{L}$, 3D surface points ($p_i$) and shape-code from $\mathcal{L}$ as input to $\mathcal{M}$, and the mapped canonical coordinate points from $\mathcal{M}$ as input to $\mathcal{C}$. We compute the $\mathcal{L}_{\text{Stage2}}$ loss using the output RGB ($r_i$), surface normals ($s_i$), and surface points ($p_i$) with the respective ground-truth $\{\{\widehat{r}_i, \widehat{s}_i, \widehat{p}_i\} \in e_i \} \in \mathcal{D}$. The learned dense correspondences are then used for TEGLO inference to extract the object texture. We show qualitative results for the correspondence maps in Fig.(\ref{fig:suppl-corr}) where canonical coordinate points from different posed images are mapped to the same location in the canonical coordinate map (canonical coordinate points are output from $\mathcal{M}$). Further, the quantitative results in Table.(\ref{tab:teglo_train}) and Table.(\ref{tab:teglo_test}) show a significant increase in PSNR and LPIPS for reconstruction - demonstrating the effectiveness of the dense correspondences learned by TEGLO Stage-2 for texture representation and tasks such as texture transfer and texture editing.

\section{Ablations}
\label{sec:suppl-ablation}
\begin{figure}[H]
  \centering
  \includegraphics[width=0.99\linewidth]{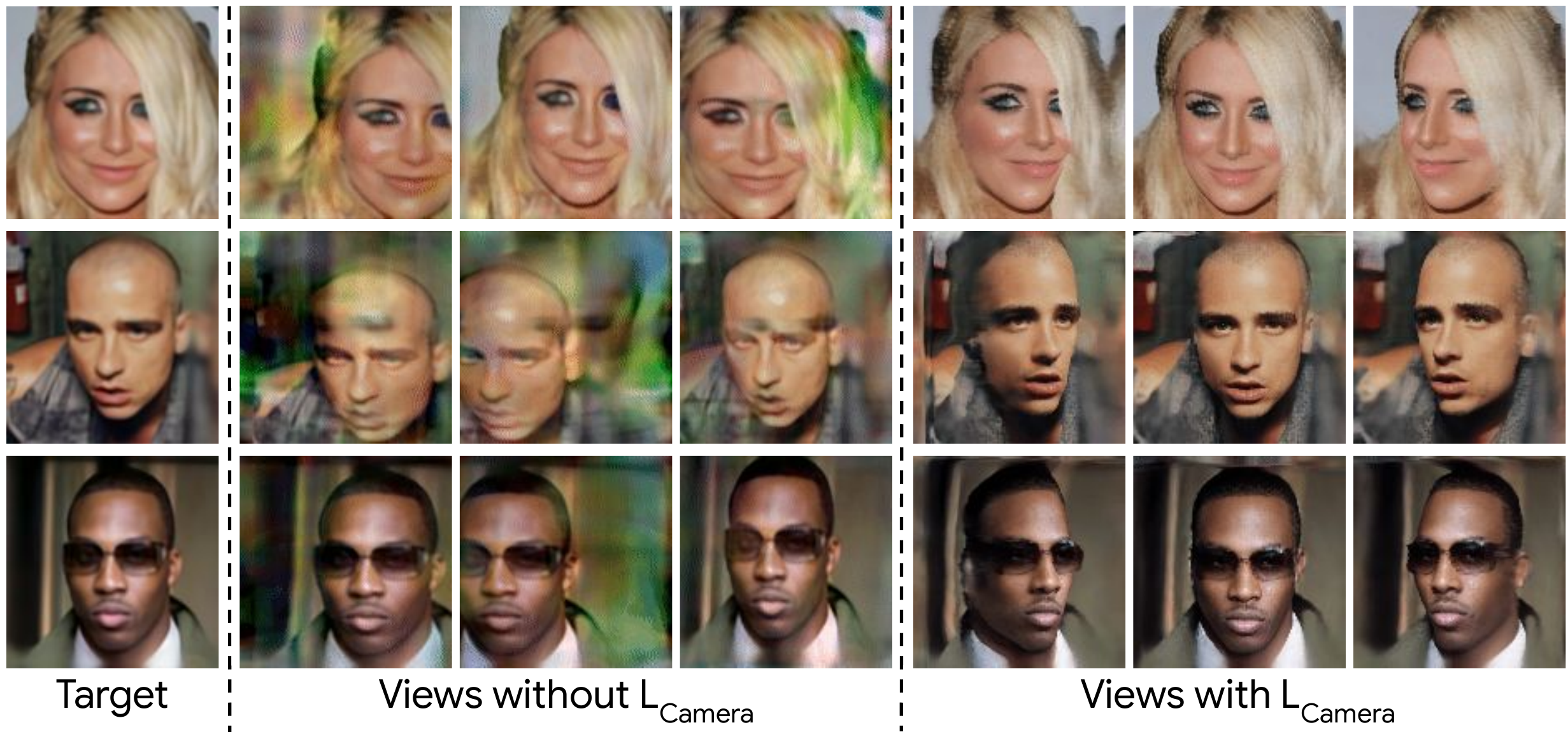}
  \caption{\textbf{Ablation} - $\mathcal{L}_{\text{Camera}}$ for TEGLO Stage-1 training.}
  \label{fig:suppl-camera}
  \vspace{-2mm}
\end{figure}

\begin{figure*}[t]
  \centering
  \includegraphics[width=0.95\linewidth]{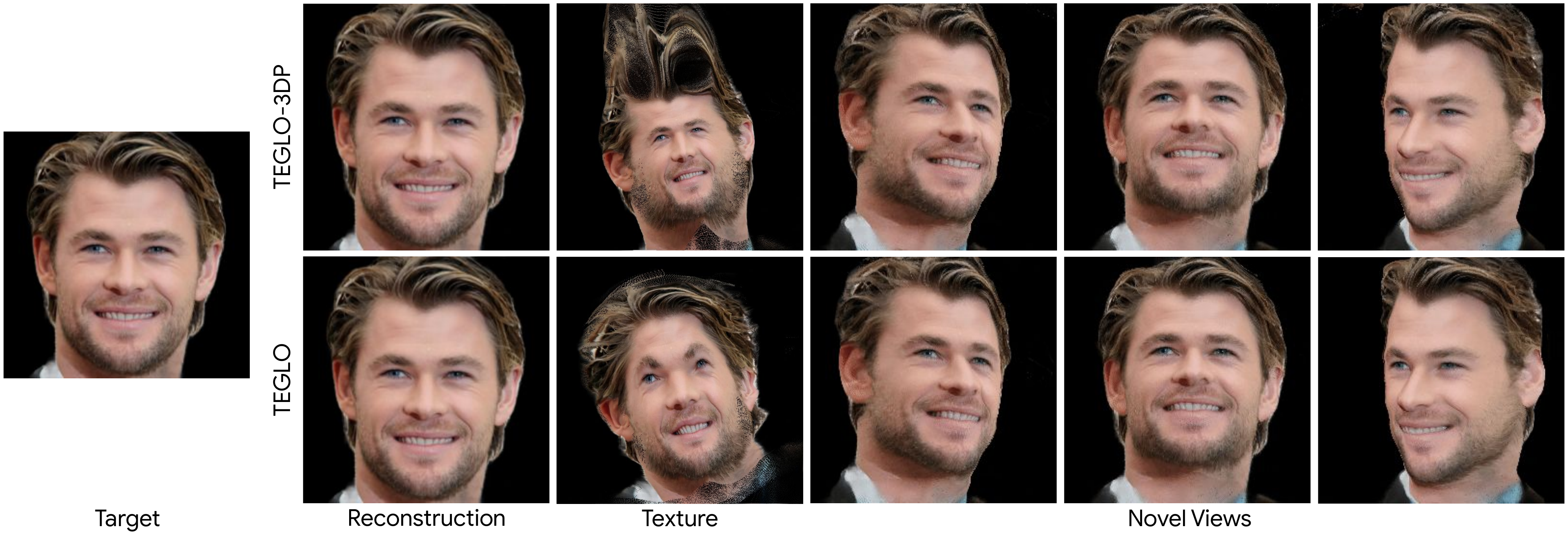}
  \caption{\textbf{Ablation} - Qualitative comparison with TEGLO-3DP (TEGLO Stage-2 with 3D surface point reconstruction only)}
  \label{fig:suppl-teglo-3dp}
\end{figure*}

\begin{figure}[H]
  \centering
  \includegraphics[width=0.99\linewidth]{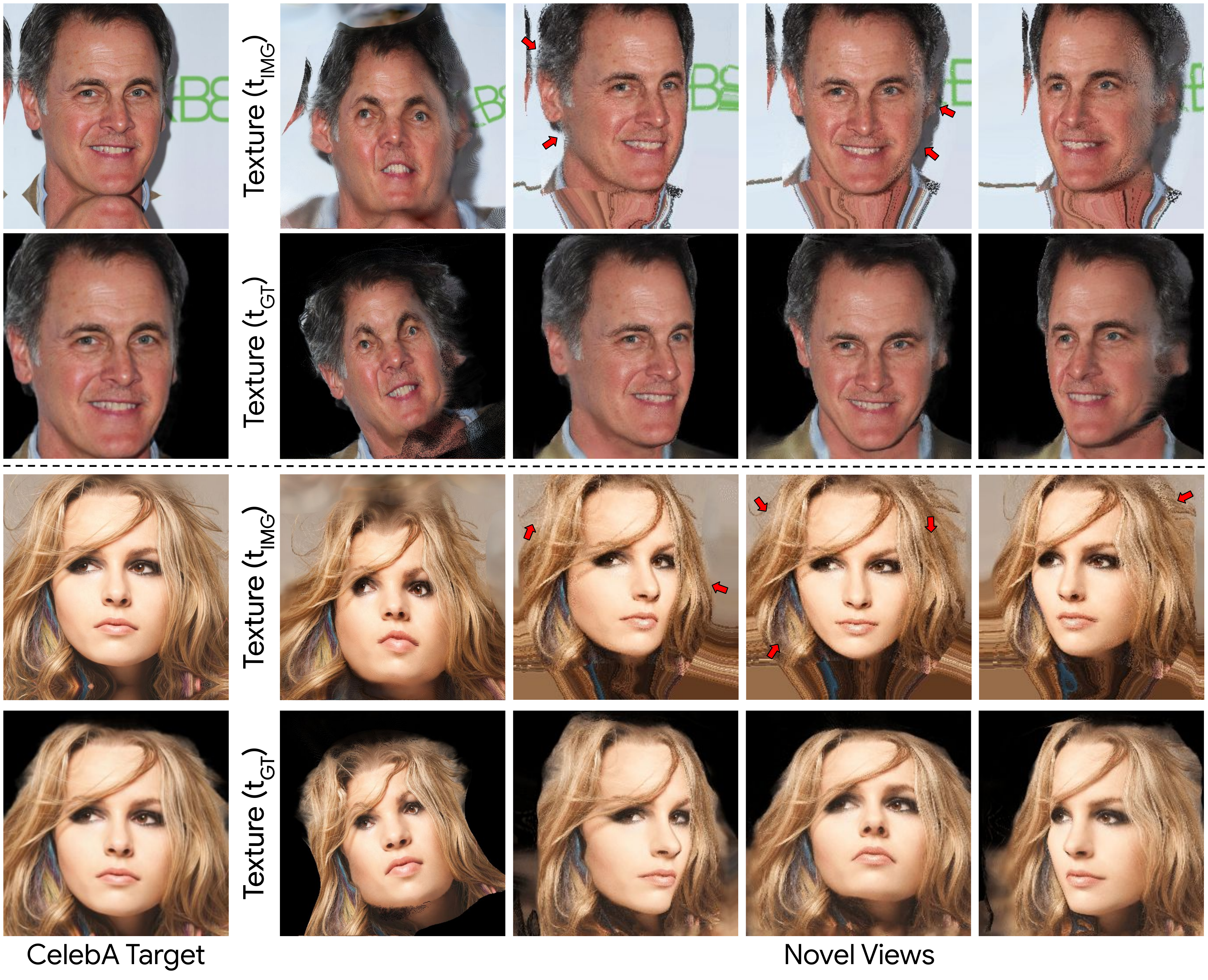}
  \caption{\textbf{Texture ablation} - Qualitative comparison with texture stored as an image indexed with bi-linear interpolation.}
  \label{fig:suppl-texture-abl}
\end{figure}

\textbf{Using $\mathcal{L}_{\text{Camera}}$ loss.} EG3D \cite{chan2022efficient} conditions the generator and discriminator with the camera pose to enable 3D consistent novel view synthesis. As noted in the main paper, the pose-conditioned generator does not completely disentangle the camera pose from appearance leading to artifacts such as facial expressions/eyes following the camera. In TEGLO Stage-1, we use the $\mathcal{L}_{\text{Camera}}, \mathcal{L}_{\text{RGB}}$ and $\mathcal{L}_{\text{Perceptual}}$ losses to train $\mathcal{N}$. An ablation experiment without the camera prediction loss led to 2D banner artifacts. This is qualitatively represented in Fig.(\ref{fig:suppl-camera}) for ``Views without $\mathcal{L}_{\text{Camera}}$" with flat and inconsistent geometries for different camera angles. However, the results for training $\mathcal{N}$ using $\mathcal{L}_{\text{Camera}}$ show multi-view consistent representations demonstrating the effectiveness of using the simple camera prediction loss. Furthermore, we show that the rendered orbits do not have expressions/eyes following the camera in the accompanying website. Qualitative results for novel view synthesis is also presented in Fig.(\ref{fig:suppl-camera}, \ref{fig:suppl-teglo-3dp}, \ref{fig:suppl-texture-abl}, \ref{fig:suppl-afhq}, \ref{fig:suppl-geo}, \ref{fig:complex-v1}, \ref{fig:suppl-sv3d}, \ref{fig:suppl-compare}, \ref{fig:suppl-textured-views}).

\textbf{K-d tree and NNI for textures.} In the main paper, we discuss the necessity of the K-d tree and Natural Neighbor Interpolation (NNI) \cite{sibson1981brief} to prevent aliasing artifacts and enable unambiguous indexing into the texture to obtain the RGB values for each surface point. The Natural Neighbor Interpolation is defined as follows:

\begin{equation}
    \mathcal{\text{NNI}}(x) = \sum_{i=0}^{n} w_i(x) \times f(x_i)
\end{equation}

\begin{equation}
    w_i(x) =  \frac{\frac{1}{d_i(x)}}{\sum\limits_{j=0}^{n} \frac{1}{d_j(x)}}
\end{equation}

Where $x$ is the query point, $w_i(x)$ is the simplified Laplace weight based on inverse distances to n neighbors corresponding to the polygon potentially encroached by the query point in the Voronoi tessellation plot, and $f(x_i)$ represents the extracted pixels from the texture. Storing the texture in the K-d tree and using Natural Neighbor Interpolation enables accurate and unambiguous floating point indexing into the texture to obtain the RGB color (this is property of NNI using a tessellation plot). We also note in the main paper about K-d tree + NNI enabling robustness to ``holes" in the texture. To verify the utility of K-d tree + NNI we design an experiment where the texture is stored as an image and indexed using bi-linear interpolation. The qualitative results are presented in Fig.(\ref{fig:suppl-texture-abl}) where $t_{\text{IMG}}$ is the texture with mapped ground truth pixels stored as an image, and $t_{\text{GT}}$ is the TEGLO texture stored as a K-d tree. We observe that the novel view synthesis using $t_{\text{IMG}}$ which includes clipping the query indices to the texture image size and bi-linear interpolation for indexing leads to several aliasing artifacts (indicated by red arrows). This is because the canonical coordinate points may not be aligned to the pixel centers and storing them in the discretized texture space may lead to imprecision which manifests as aliasing artifacts in novel view synthesis. We further note that the quantitative results for test set reconstruction using $t_{\text{IMG}}$ is 30.828 dB PSNR and 0.0823 for LPIPS at $256^2$ resolution for CelebA-HQ data. In contrast, novel view synthesis using $t_{\text{GT}}$ leads to a significant reduction in aliasing artifacts with quantitative results of 86.2 dB PSNR and 7.4e-7 LPIPS for test set reconstruction of CelebA-HQ images at $256^2$ resolution (refer  Fig.(\ref{fig:suppl-texture-abl}) and the accompanying website). 

\textbf{TEGLO Stage-2 with $\mathcal{L}_{\text{Coord}}$ loss only.} Previous work AUV-Net \cite{chen2022auv} states that methods \cite{deng2021deformed, liu2020learning} that do not use color for learning dense correspondences may learn sub-par texture representations. To verify, we train TEGLO Stage-2 with only $\mathcal{L}_{\text{Coord}}$ reconstruction loss instead of $\mathcal{L}_{\text{Stage2}} = \mathcal{L}_{\text{Coord}} + \mathcal{L}_{\text{Coord}} + \mathcal{L}_{\text{Coord}}$ reconstruction losses. The qualitative results are presented in Fig.(\ref{fig:suppl-compare}) comparing TEGLO-3DP with TEGLO and other baseline results. Of particular interest is Fig.(\ref{fig:suppl-teglo-3dp}) with qualitative results for TEGLO and TEGLO-3DP including the texture image. We note that the reconstruction and novel view synthesis results are nearly identical. However, we also observe TEGLO-3DP including a wayward texture representation near the hair region. While the dense correspondences map the surface points to the appropriate RGB image pixels, there is a scope for null pixel artifacts around the hair region when using NNI. While the 3D reconstruction and novel view synthesis for TEGLO-3DP and TEGLO do not differ, we note the potential for black pixels to be obtained in novel view synthesis leading to lowered qualitative and quantitative results.

\textbf{Limitations.} As mentioned in the main paper, training TEGLO involves a computational overhead. Further, TEGLO is only able to map target image pixels spanning the target image and hence there are missing pixel artifacts for camera views with minimal mapped target image pixels. For example, the texture ($T_{\text{GT}}$) in row-4 in Fig.(\ref{fig:suppl-textured-views}) includes missing pixels near the lower jaw region. In the last column in row-4, the novel rendered view does not include any target image pixels for the ear. On a similar note, in row-2, column-7 in Fig.(\ref{fig:complex-v1}), the novel view shows a slight twist in the nose geometry partially due to the thin veil on the face which could not be accounted for by TEGLO Stage-1. 

\vspace{-1mm}
\section{Qualitative results}
\label{sec:suppl-qual}
\begin{figure}[H]
  \centering
  \scalebox{0.99}{
  \includegraphics[width=0.99\linewidth]{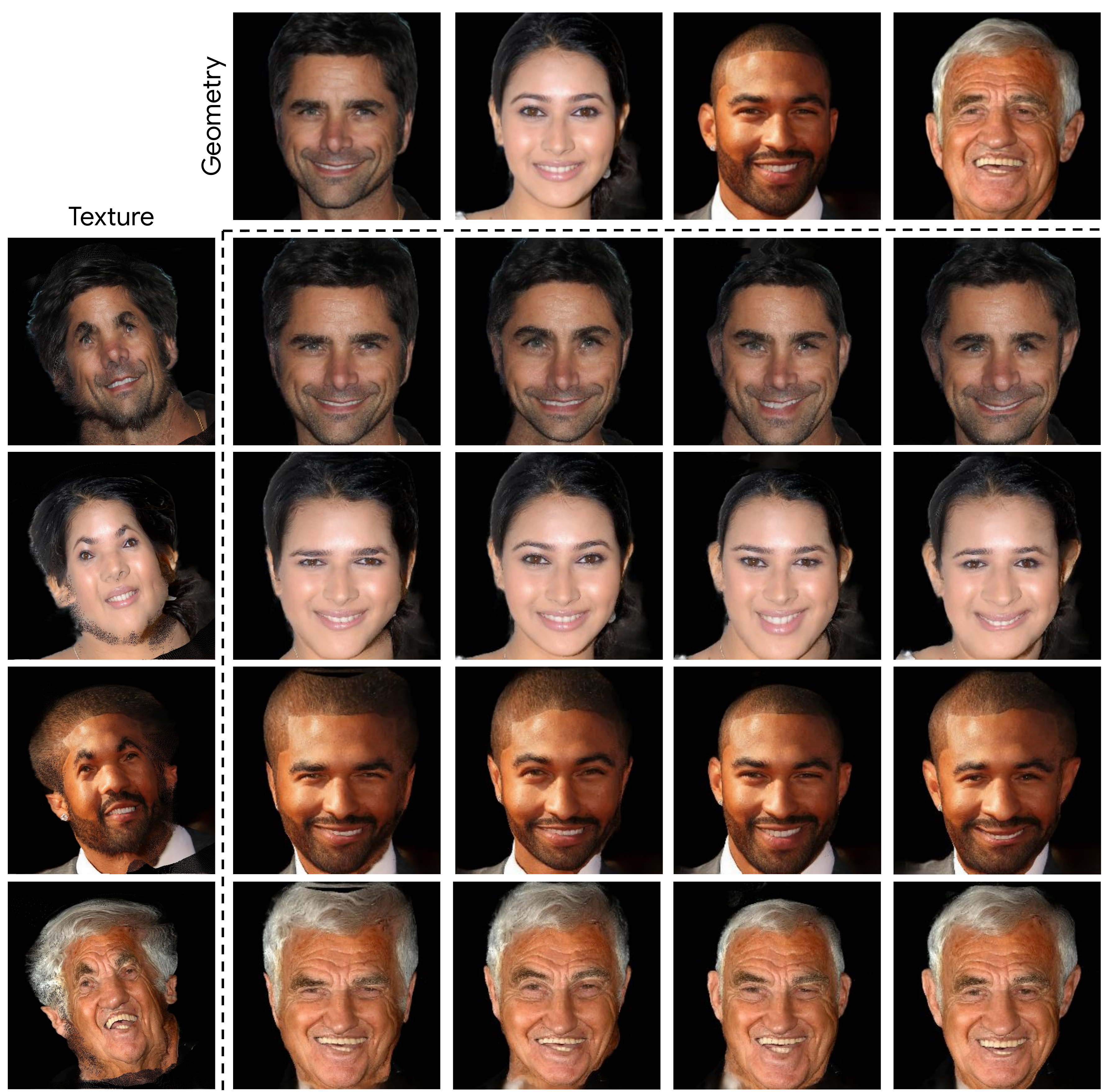}
  }
  \caption{\textbf{Texture transfer} - Qualitative results for texture transfer with CelebA-HQ. (Top row shows CelebA-HQ image targets).}
  \label{fig:suppl-transfer}
\end{figure}

\begin{figure}[H]
\centering
\scalebox{0.99}{
  \includegraphics[width=0.99\linewidth]{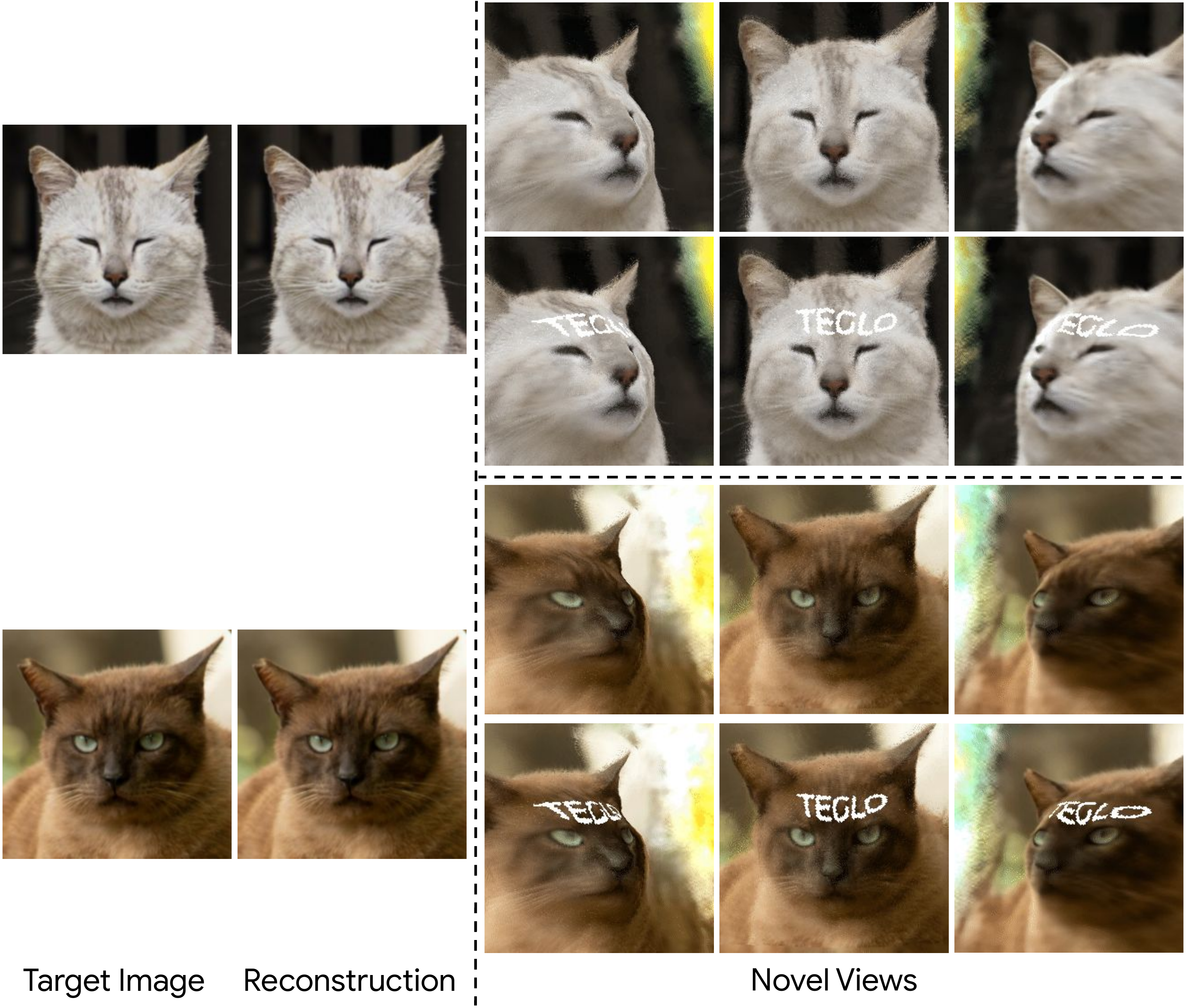}
  }
  \caption{\textbf{Textured synthesis} - Target view reconstruction and novel view synthesis for AFHQv2-Cats.}
  \label{fig:suppl-afhq}
  \vspace{-2mm}
\end{figure}

\begin{figure}[H]
  \centering
\scalebox{0.99}{
  \includegraphics[width=0.99\linewidth]{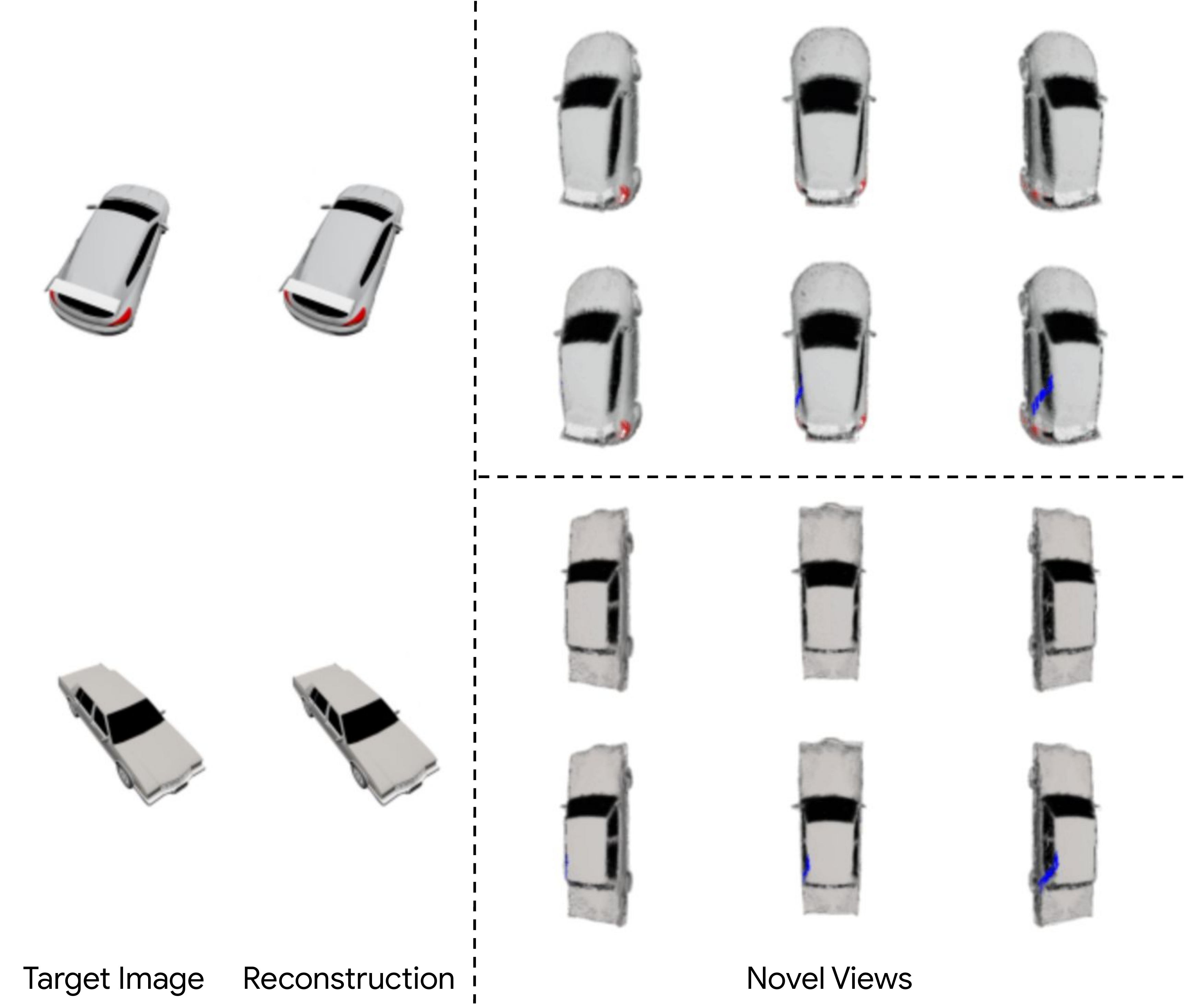}
  }
  \caption{\textbf{Textured synthesis} - Target view reconstruction and novel view synthesis for SRN-Cars.}
  \label{fig:suppl-cars}
  \vspace{-2mm}
\end{figure}

\begin{figure}[H]
  \centering
\scalebox{0.9}{
  \includegraphics[width=0.99\linewidth]{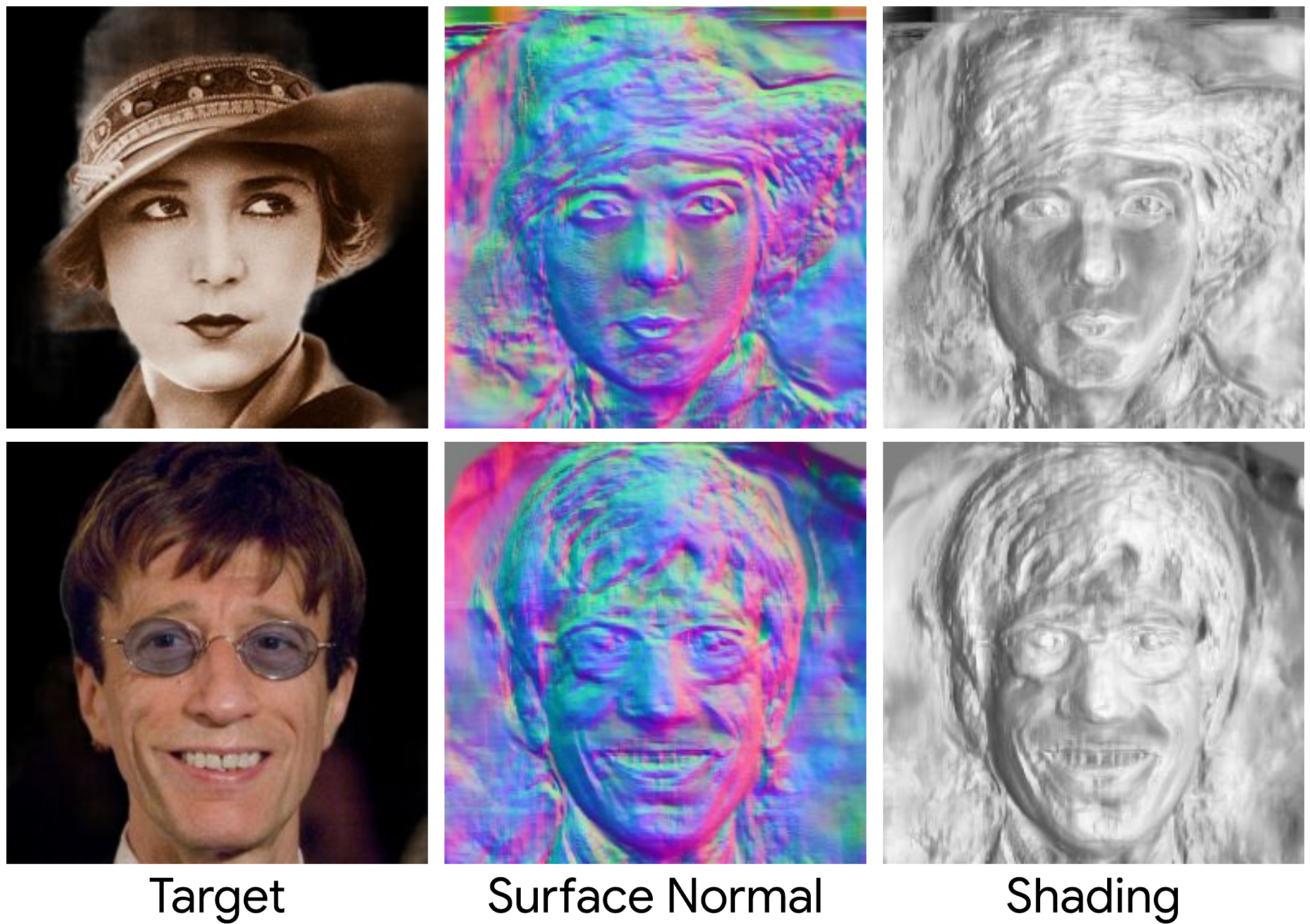}
  }
  \caption{\textbf{Geometry results} - Target view reconstruction geometry for CelebA-HQ with hat and eyeglasses.}
  \label{fig:suppl-geo}
  \vspace{-2mm}
\end{figure}

\begin{figure*}[t]
  \centering
  \includegraphics[width=0.99\linewidth]{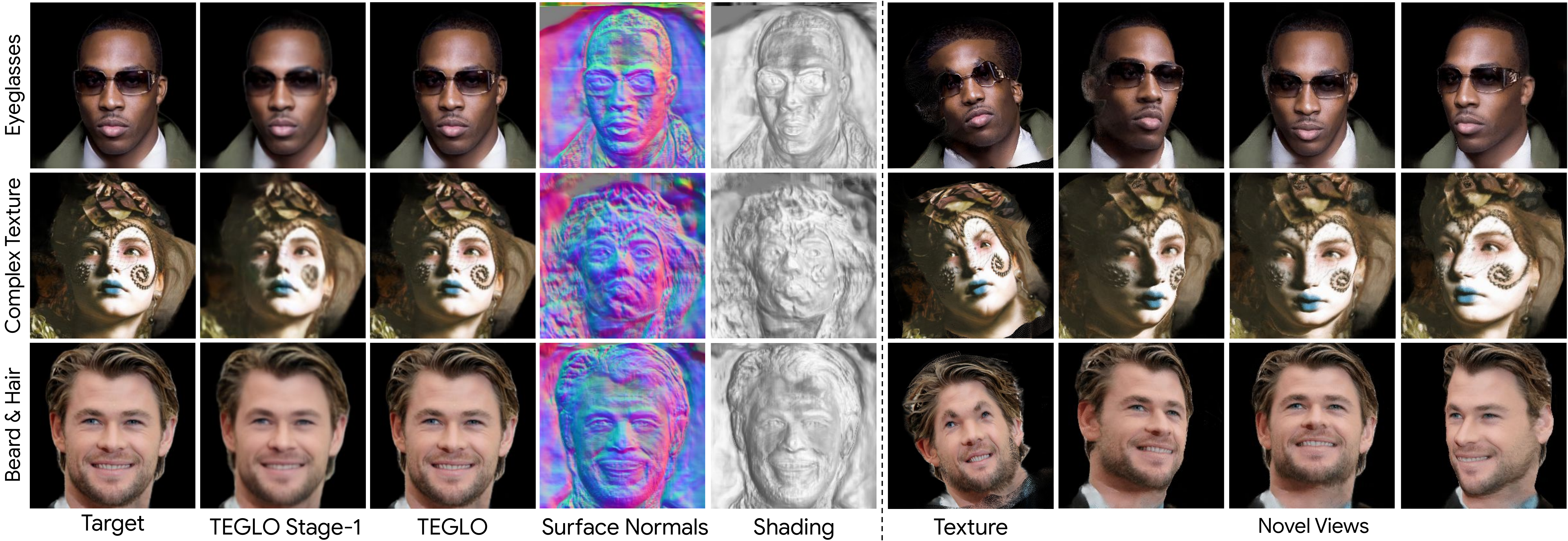}
  \caption{\textbf{Complex geometry and texture} - Qualitative results for texture edits.}
  \label{fig:complex-v1}
\end{figure*}

\begin{figure*}[t]
  \centering
\scalebox{0.99}{
  \includegraphics[width=0.99\linewidth]{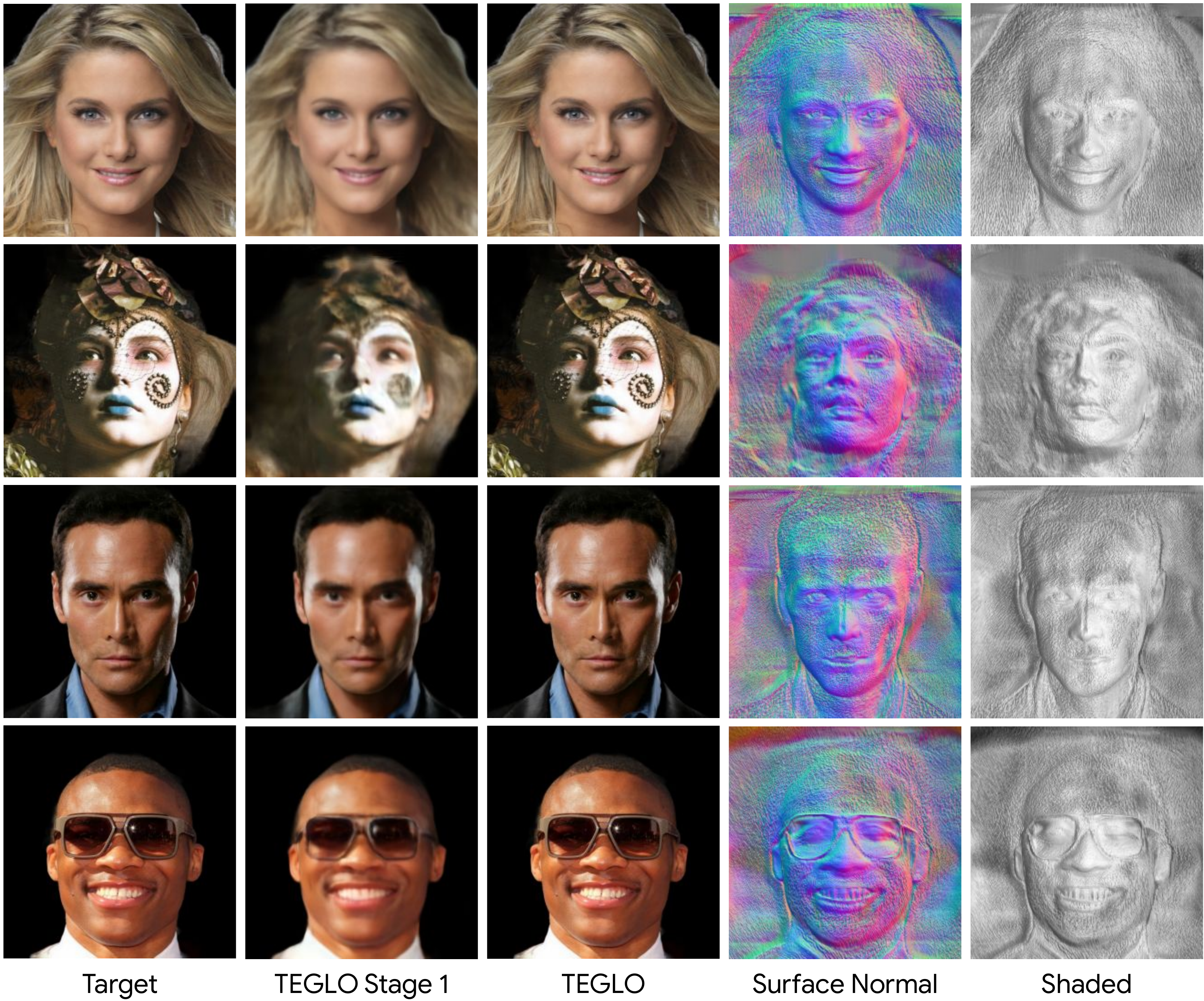}
  }
  \caption{\textbf{Single view 3D reconstruction} - Results for TEGLO trained on FFHQ data and evaluated on CelebA-HQ image targets.}
  \label{fig:suppl-sv3d}
\end{figure*}

\begin{figure*}[t]
  \centering
  \includegraphics[width=0.99\linewidth]{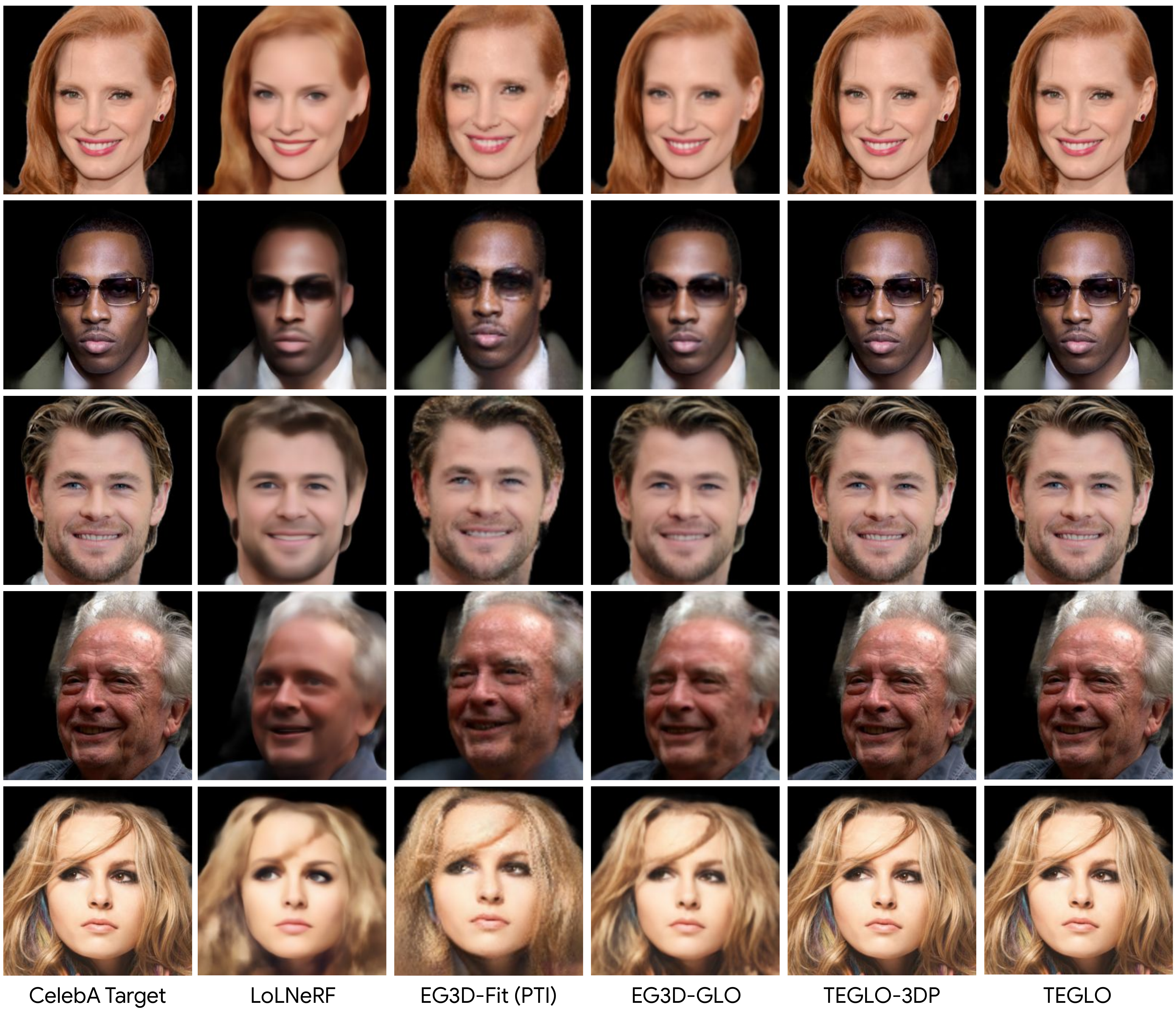}
  \caption{\textbf{Qualitative results} - Comparison with relevant 3D-aware generative baseline methods at $256^2$ resolution for CelebA-HQ.}
  \label{fig:suppl-compare}
\end{figure*}

\begin{figure*}[t]
  \centering
  \includegraphics[width=0.916\linewidth]{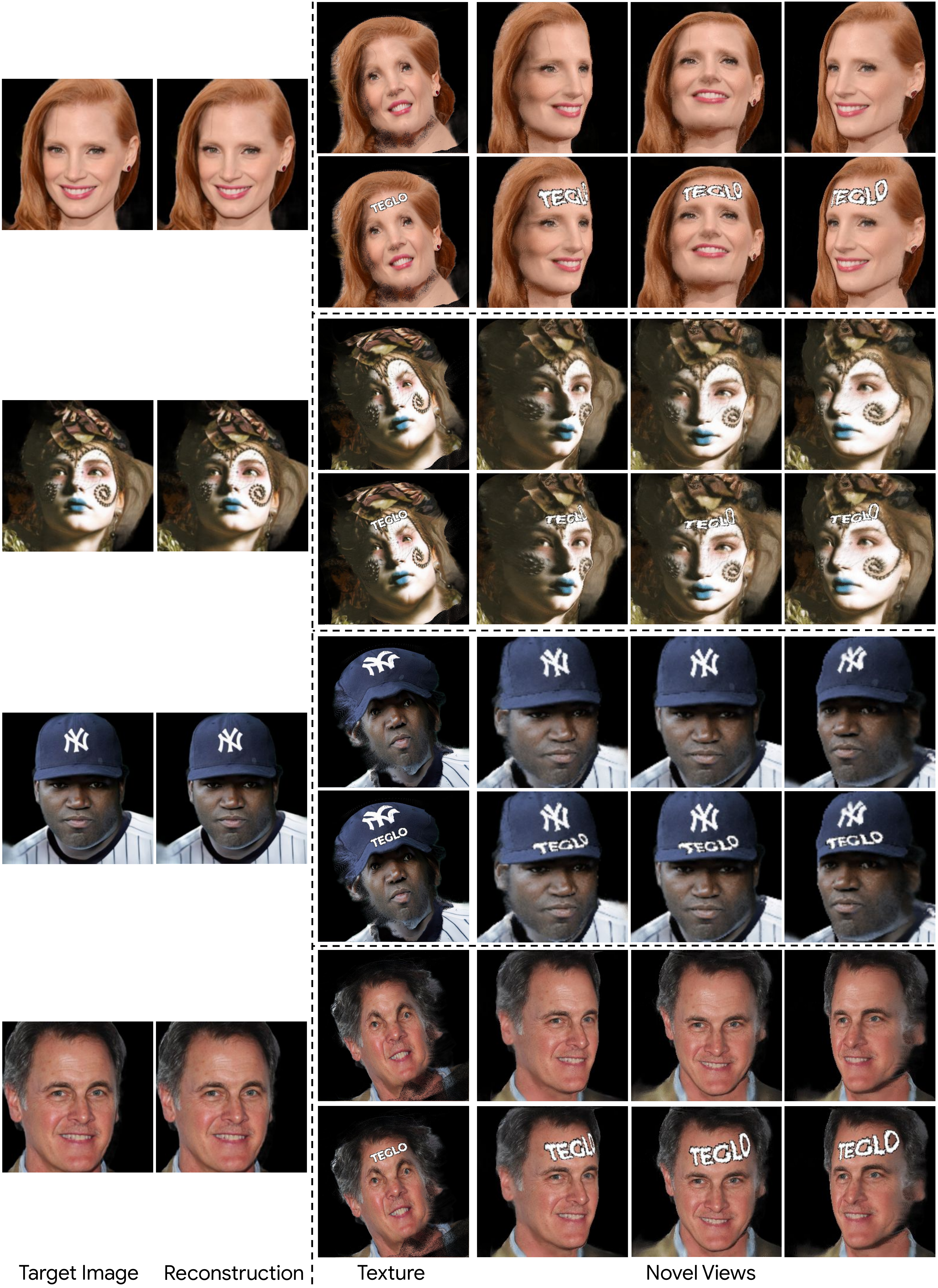}
  \caption{\textbf{Textured synthesis} - Target view reconstruction and novel view synthesis for CelebA-HQ.}
  \label{fig:suppl-textured-views}
\end{figure*}

We present qualitative results for texture transfer in Fig.(\ref{fig:suppl-transfer}). We present qualitative results for texture edits and textured synthesis for AFHQv2-Cats in Fig.(\ref{fig:suppl-afhq}), for SRN-Cars in Fig.(\ref{fig:suppl-cars}) and for CelebA-HQ in Fig.(\ref{fig:suppl-textured-views}). We present single-view textured 3D reconstruction results for TEGLO trained on FFHQ data and evaluated on CelebA-HQ image targets - focus on complex details such eyeglasses and make-up - in Fig.(\ref{fig:suppl-sv3d}). We present extended results for Fig.(\ref{fig:complex}) in Fig.(\ref{fig:complex-v1}) to compare row-1, row-2 with Fig.(24) in \cite{cao2022authentic} and row-3 with Fig.(25) in \cite{li2020dynamic}). We also include surface normal and shading results with hats and eyeglasses in Fig.(\ref{fig:suppl-geo}). Further, we also present comparative qualitative results with baselines and TEGLO-3DP (TEGLO Stage-2 trained with only $\mathcal{L}_{\text{Coord}}$ loss) in Fig.(\ref{fig:suppl-compare}). Lastly, we present qualitative results for high-resolution rendering at $1024^2$ resolution in Fig(\ref{fig:high-res}) and show novel view synthesis with a specific focus on high-frequency details such as freckles (pigments under the skin) in row-1, make-up and jewelry in row-2, hair and beard in row-3, fine skin details in row-4 and wrinkles in row-5.

\section{Efficient dataset rendering} 
\begin{figure*}[t]
  \centering
  \includegraphics[width=0.97\linewidth]{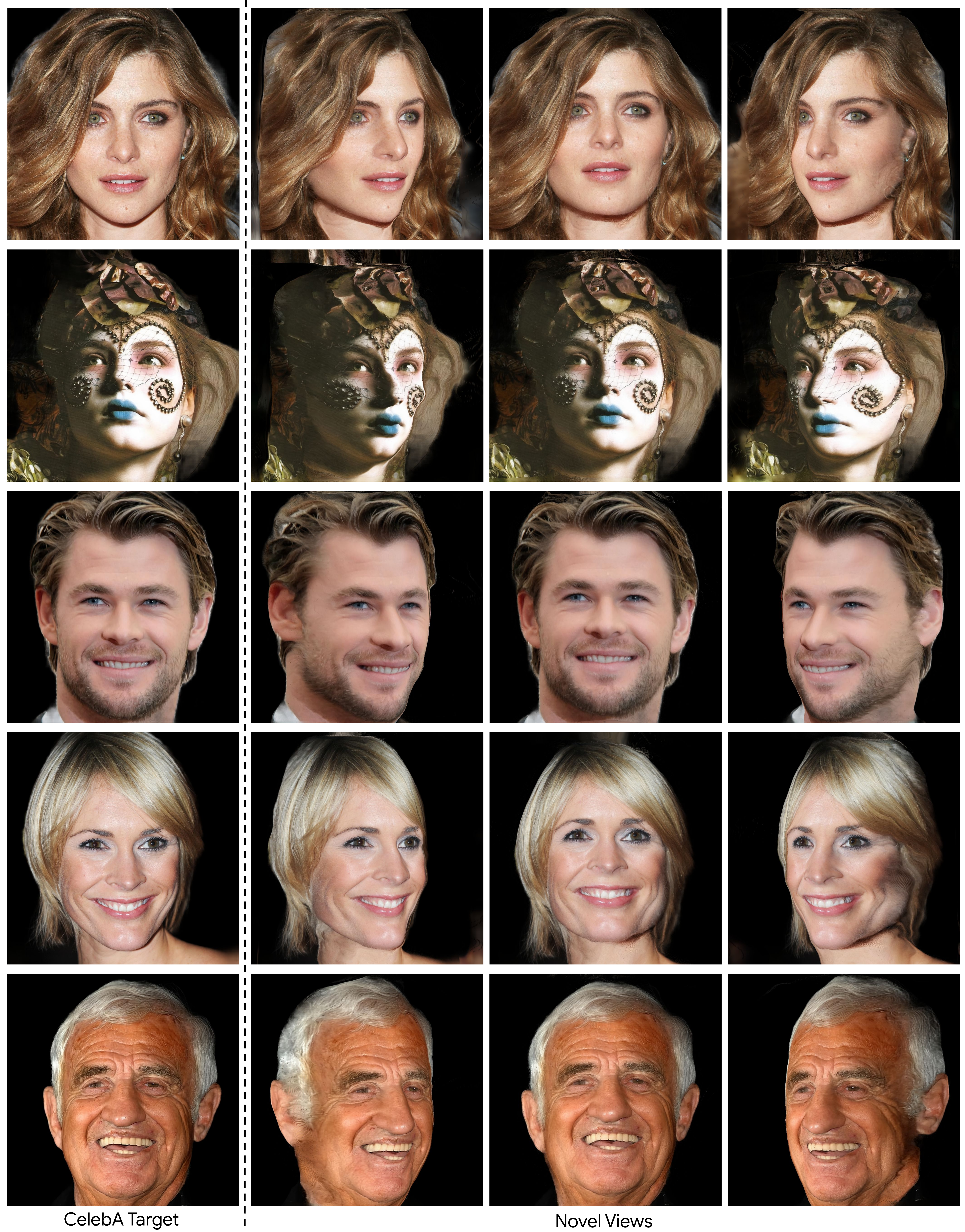}
  \caption{\textbf{High resolution rendered views} - Qualitative results for novel views in high-resolution with high-frequency details such as freckles, jewelry, make-up, hair, fine details and wrinkles.}
  \label{fig:high-res}
\end{figure*}

TEGLO enables 3D high-fidelity data rendering from single-view image collections of objects at arbitrary resolution. Since the dense correspondences are learned point-wise (\ie using $\widehat{p}_j$), there is no spatial constraint in querying $\mathcal{M}$ for the canonical coordinate point. Hence, we render images of any size by first dividing the image pixels into 4 tiles, then obtain the surface points from TEGLO Stage-1, map the surface points to the canonical coordinate space using $\mathcal{M}$ and then index into the texture to obtain the RGB color value. Then, after all the tiles are computed, we can combine the divided computations into a single image of high-resolution. The orbit and the high-resolution frames at $1024^2$ resolution in Fig.(\ref{fig:high-res}) and on the website have been obtained using this approach.

\vspace{-1mm}
\section{Ethical considerations}
\label{sec:suppl-ethical}
One of the motivating goals for TEGLO stems from the need for photorealistic 3D reconstruction of objects from single-view image collections. By using a compiled set of images, TEGLO helps mitigate bias, reduce costs and also improve access to high-quality data for the broader research community. Hence, TEGLO enables rendering a dataset of diverse objects (improving fairness and mitigating bias) and also reduces the need for large scale data collection. Further, we acknowledge the potential misuse of TEGLO especially for single view 3D reconstruction and hence only make available the code for reproducibility purposes. We will not release any trained weights for our method.



\end{multicols}

\end{document}

%% file: Sections/1_Intro.tex
\begin{figure}[t]
  \centering
  \includegraphics[width=0.99\linewidth]{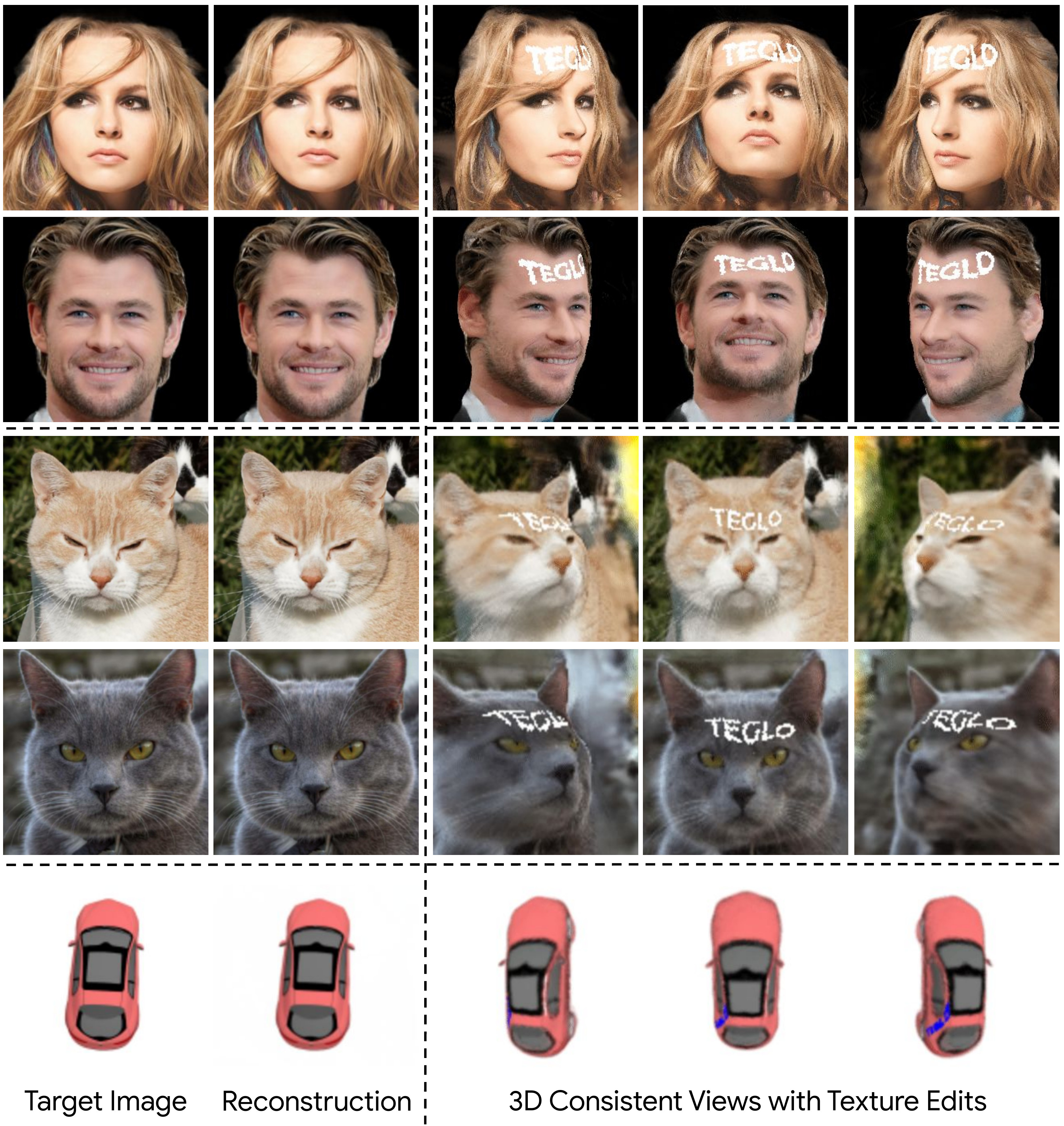}
  \caption{\textbf{Teaser} - Demonstrating TEGLO for high fidelity 3D reconstruction and multi-view consistent texture representation and texture editing from single-view image collections of objects.}
  \label{fig:teaser}
  \vspace{-5mm}
\end{figure}

Reconstructing high-resolution and high-fidelity 3D consistent representations from single-view in-the-wild image collections is critical for applications in virtual reality, 3D content creation and telepresence systems. Recent work in Neural Radiance Fields (NeRFs) \cite{chan2021pi, gu2021stylenerf, chan2022efficient, rebain2022lolnerf} aim to address this by leveraging the inductive bias across a dataset of single-view images of class-specific objects for 3D consistent rendering. However, they are unable to preserve high frequency details while reconstructing the input data despite the use of SIREN \cite{sitzmann2020implicit} or positional encoding \cite{mildenhall2021nerf}, in part due to the properties of MLPs that they use \cite{chen2022auv}. For arbitrary resolution 3D reconstruction from single-view images, these methods face several challenges such as image-space approximations that break multi-view consistency constraining the rendering resolution \cite{chan2022efficient}, requiring Pivotal Tuning Inversion (PTI) \cite{roich2022pivotal} or fine-tuning for reconstruction \cite{gu2021stylenerf, chan2022efficient, skorokhodov2022epigraf} and the inability to preserve high-frequency details \cite{gu2021stylenerf, chan2022efficient, skorokhodov2022epigraf, rebain2022lolnerf}. To address these limitations, we propose TEGLO (Textured EG3D-GLO) that uses a tri-plane representation ~\cite{chan2022efficient} and Generative Latent Optimization (GLO) \cite{bojanowski2017optimizing} based training to enable efficient and high-fidelity 3D reconstruction, and novel view synthesis at arbitrary image resolutions from single-view image collections of objects.


\begin{figure}[t]
  \centering
  \scalebox{0.9}{
  \includegraphics[width=0.9\linewidth]{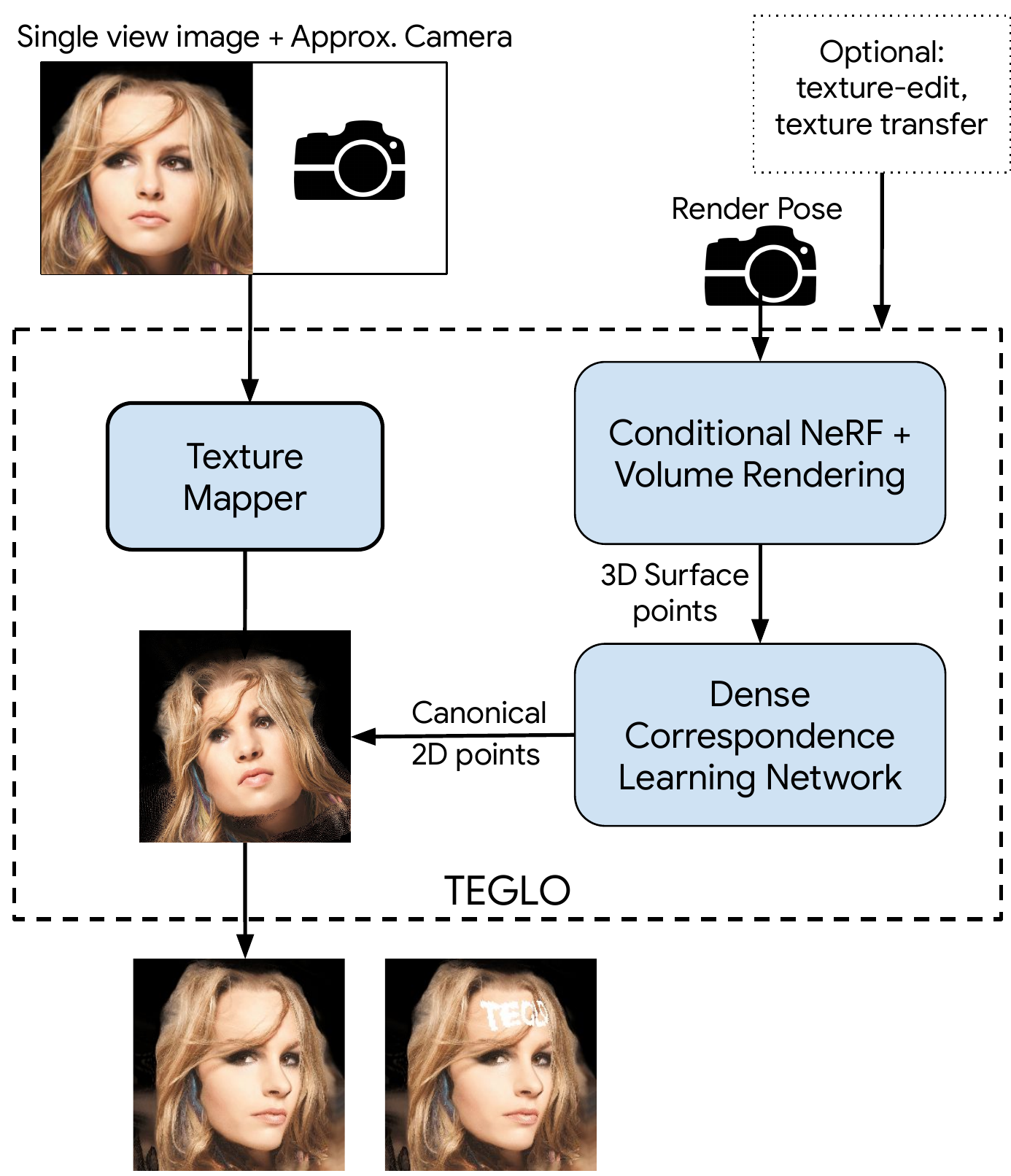}
  }
  \caption{\textbf{Overview} - TEGLO enables 3D reconstruction and texture representation from single-view image collections of objects.}
  \label{fig:teglo-overiew}
  \vspace{-3mm}
\end{figure}

Recent works disentangle texture from geometry \cite{chen2022auv, yang2022neumesh} and enable challenging tasks such as texture editing and texture transfer. However, they depend on large-scale textured mesh data for high-fidelity 3D reconstruction which is laborious, expensive and time intensive to capture. Further, the use of a capture environment may cause a dataset-shift leading to generalization issues in downstream tasks, and the data use may require custom licensing. All of these factors limit access from the broader research community. This motivates the need for a method to learn textured 3D representations from single-view in-the-wild images of objects. However, the task of disentangling texture and 3D geometry from in-the-wild image collections is a formidable challenge due to the presence of wide variations in poses, partial views, complex details in appearance, geometry, noise \etc in the given image collection. Inspired by surface fields \cite{goli2022nerf2nerf}, TEGLO leverages the 3D surface points of objects extracted from a NeRF to learn dense correspondences via a canonical coordinate space to enable texture transfer, texture editing and high-fidelity single-view 3D reconstruction.



Our key insight is that by disentangling texture and geometry using the 3D surface points of objects to learn a dense correspondence mapping via a 2D canonical coordinate space, we can extract a texture for each object. Then, by using the learned correspondences to map the pixels from the input image of the object onto the texture, we enable preserving high-frequency details. As expected, copying the input image pixels onto the texture accurately allows near perfect reconstruction while preserving high-fidelity multi-view consistent representation with high-frequency details. In this work, we present TEGLO, consisting of a tri-plane and GLO-based conditional NeRF and a method to learn dense correspondences to enable challenging tasks such as texture transfer, texture editing and high-fidelity 3D reconstruction even at large megapixel resolutions. We also show that TEGLO enables single-view 3D reconstruction with no constraints on resolution by inverting the image into the latent table without requiring PTI \cite{roich2022pivotal} or model fine-tuning. We present an overview of our final model in Fig.(\ref{fig:teglo-overiew}): TEGLO takes a single-view image and its approximate camera pose to map the pixels onto a texture. Then, to render the object from a different view, we extract the 3D surface points from the trained NeRF and use the dense correspondences to obtain the color for each pixel from the mapped canonical texture. Optionally, TEGLO can take texture edits and transfer textures across objects. In summary, our contributions are:

\begin{enumerate}[noitemsep,topsep=1pt,itemsep=1pt]
    

    
    \item A framework for effectively mapping the pixels from an in-the-wild single-view image onto a texture to enable high-fidelity 3D consistent representations preserving high-frequency details.
    
    
    
    
    \item A method for extracting canonical textures from single-view images enabling tasks such as texture editing and texture transfer for NeRFs.
    
    \item Demonstrating that we can effectively map the single-view image pixels to canonical texture space while preserving 3D consistency and so achieving near perfect reconstruction ($\geq 74$ dB PSNR at $1024^2$ resolution). 

\end{enumerate}



%% file: Sections/2_Related.tex
\begin{figure*}[t]
  \centering
  \includegraphics[width=0.95\linewidth]{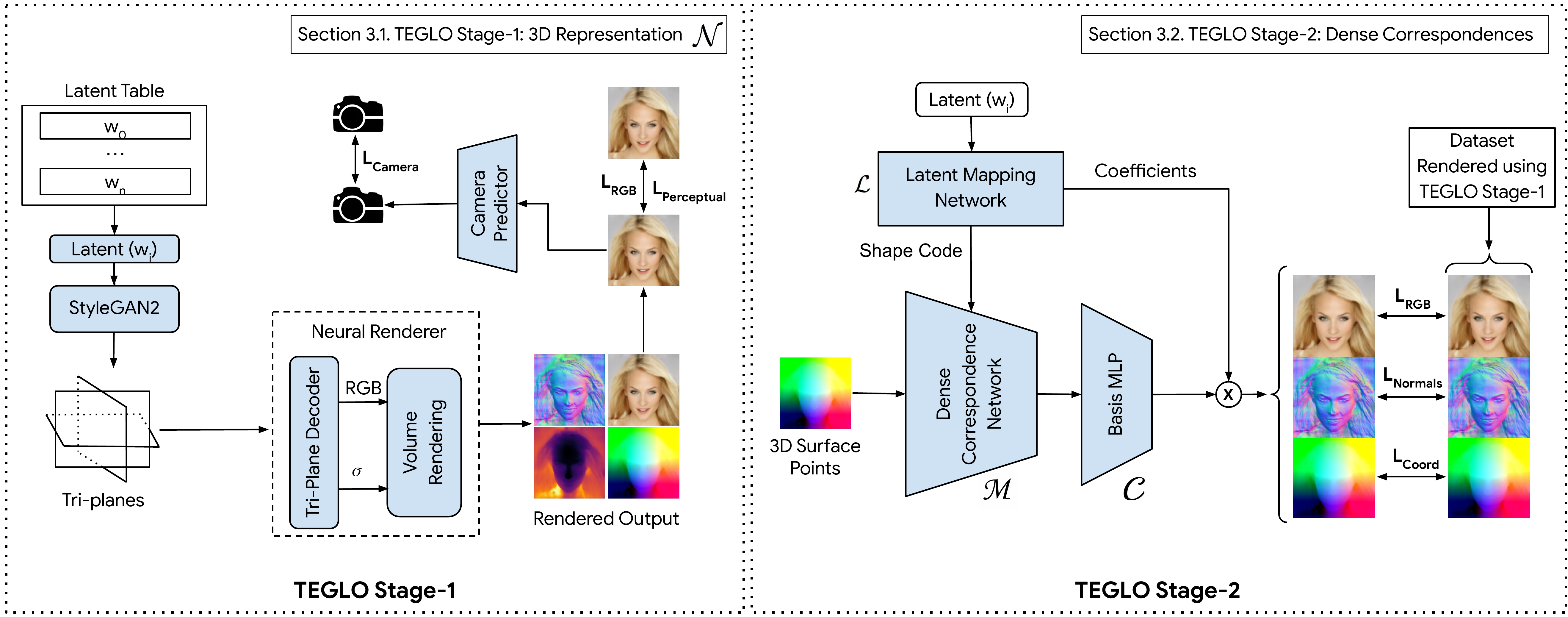}
  \caption{\textbf{Architecture} - TEGLO Stage-1 (left) uses a tri-plane and GLO based conditional NeRF to learn a per-object table of latents to reconstruct the single-view image collection. TEGLO Stage-2 (right) learns dense correspondences via a 2D canonical coordinate space.}
  \label{fig:eg3d_glo}
  \vspace{-4mm}
\end{figure*}

\textbf{3D-aware generative models.} Learning 3D representations from multi-view images with camera poses have been extensively studied since the explosion of Neural Radiance Fields (NeRFs) \cite{mildenhall2021nerf, srinivasan2021nerv, zhang2020nerf++, barron2021mip, zhang2021nerfactor, gu2021stylenerf}. However, these methods require several views and learn a radiance field for a single scene. RegNeRF \cite{niemeyer2022regnerf} reduces the need from several views to only a handful, however, the results have several artifacts. Recently, several works learn 3D representations from single-view images \cite{chan2021pi, chan2022efficient, lin20223d, skorokhodov2022epigraf, rebain2022lolnerf, zhao2022generative}. Further, \cite{sun2022fenerf, sun2022ide, tang2022explicitly, kim20233d} enable multi-view consistent editing, however, they are limited by the rendering resolution. Recent work propose single image 3D consistent novel view synthesis \cite{yu2021pixelnerf, lin2023vision, gu2023nerfdiff, watson2022novel}, however they are not yet suitable for texture representation. While point cloud based diffusion models \cite{zeng2022lion, nichol2022point} enable learning 3D representations, they have limited applicability in textured 3D generation and high fidelity novel view synthesis. In this work, we show that TEGLO learns textured 3D representations from class-specific single-view image collections.


\textbf{Texture representation.} Template based methods \cite{pavllo2021learning, bhattad2021view, chen2019learning, henderson2020leveraging} deform a template mesh prior for 3D representations and are hence restricted in the topology they can represent. Texture Fields \cite{oechsle2019texture} enable predicting textured 3D models given an image and a 3D shape, but are unable to represent high-frequency details. While NeuTex \cite{xiang2021neutex} enables texture representation, it does not allow multi-view consistent texture editing at the desired locations due to a contorted UV mapping \cite{yang2022neumesh}. NeuMesh \cite{yang2022neumesh} learns mesh representations to enable texture transfer and texture editing using textured meshes. However, it performs mesh-guided texture transfer and requires spatial-aware fine-tuning for mesh-guided texture edits. While GET3D \cite{gao2022get3d} learns textured 3D shapes by leveraging tri-plane based geometry and texture generators, it requires 2D silhouette supervision and is limited to synthetic data. AUVNet \cite{chen2022auv} represents textures from textured meshes by learning an aligned UV mapping and demonstrates texture transfer. However, it depends on textured mesh data and requires multiple networks to enable single-view 3D reconstruction. In contrast, TEGLO learns textured 3D consistent representations from single-view images by inverting the image into the latent table.



\textbf{Dense correspondences.} Previous work in dense correspondence learning involve supervised \cite{choy2016universal, li2019joint} or unsupervised \cite{xu2021rethinking, xiao2022learning} learning methods. CoordGAN \cite{mu2022coordgan} learns dense correspondences by extracting each image as warped coordinate frames transformed from correspondence maps which is effective for 2D images. However, CoordGAN is unable to learn 3D correspondences. AUVNet \cite{chen2022auv} establishes dense correspondences across 3D meshes via a canonical UV mapping and asserts that methods that do not utilize color for dense correspondence learning \cite{deng2021deformed, liu2020learning} may have sub-par performance in texture representation.

%% file: Sections/3_Methods.tex
\vspace{-1mm}


Given a collection of single-view in-the-wild images of objects and their approximate camera poses, TEGLO aims to learn a textured 3D representation of the data. TEGLO consists of two stages: 3D representation learning and dense correspondence learning. TEGLO Stage-1 consists of a conditional NeRF leveraging a Tri-Plane representation and an auto-decoder training regime based on generative latent optimization (GLO) \cite{bojanowski2017optimizing} for 3D reconstruction of the image collection. TEGLO Stage-2 uses a dataset rendered using TEGLO Stage-1 consisting of the geometry from five views of an object and the optimized latent code. TEGLO Stage-2 uses the 3D surface points from the rendered dataset to learn dense pixel-level correspondences via a 2D canonical coordinate space. Then, the inference stage uses the learned dense correspondences to map the image pixels from the single-view input image onto a texture extracted from TEGLO-Stage 2. As a result, TEGLO effectively preserves high frequency details at an unprecedented level of accuracy even at large megapixel resolutions. TEGLO disentangles texture and geometry enabling texture transfer (Fig.(\ref{fig:texture-transfer})), texture editing (Fig.(\ref{fig:editing})) and single view 3D reconstruction without requiring fine-tuning or PTI (Fig.(\ref{fig:single-view-3d})). 



\subsection{TEGLO Stage 1: 3D representation}
\label{sec:data}
\textbf{Formulation.} We denote the single-view image collection ($\mathcal{I}$) with class specific objects as $\{o_0, o_1, ..., o_n\} \in \mathcal{I}$. For learning 3D representations, TEGLO employs a generative latent optimization (GLO) based auto-decoder training, where NeRF is conditioned on an image specific latent code $\{w_0, w_1, ..., w_n\} \in \mathcal{R}^D$ to effectively reconstruct the image without requiring a discriminator.

\begin{figure}[t]
  \centering
  \includegraphics[width=0.99\linewidth]{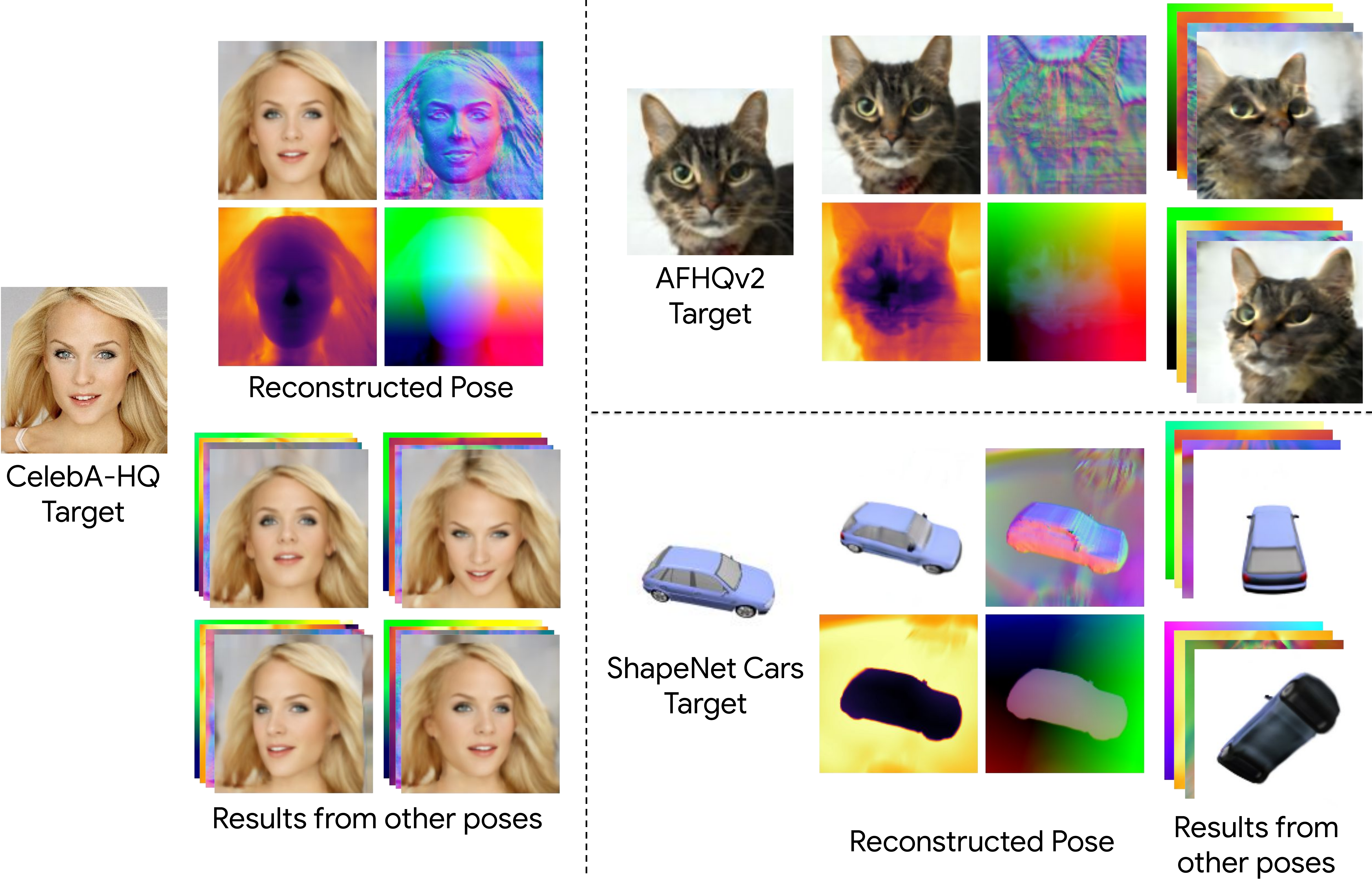}
  \caption{\textbf{Rendering the dataset for TEGLO Stage-2} - Rendering multiple views of images, surface normals, depth maps and 3D surface points from CelebA-HQ, AFHQv2-Cats and ShapeNet-Cars for learning dense correspondences in TEGLO Stage-2.}
  \label{fig:eg3dglo_data}
  \vspace{-3mm}
\end{figure}

\begin{figure}[t]
  \centering
  \scalebox{0.99}{
  \includegraphics[width=0.99\linewidth]{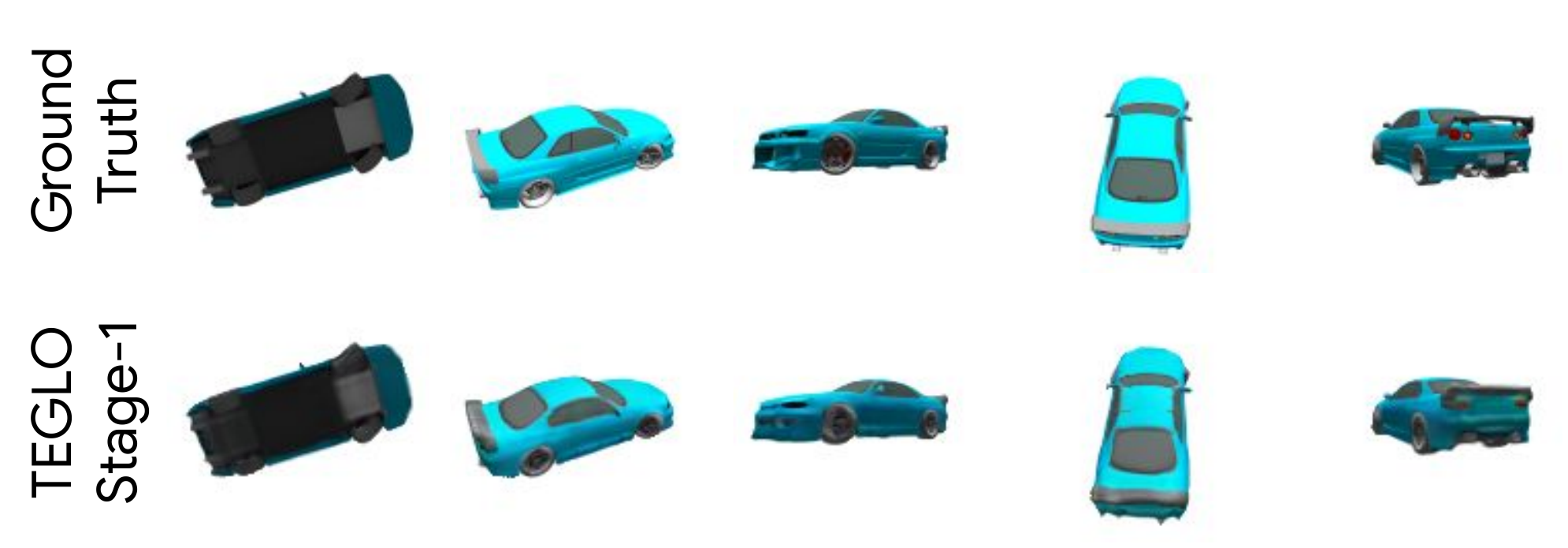}
  }
  \caption{\textbf{Novel view synthesis} - Results for ShapeNet-Cars data.}
  \label{fig:nvs}
  \vspace{-2mm}
\end{figure}

\textbf{Network architecture.} The NeRF model $\mathcal{N}$ is represented by TEGLO Stage-1 in Fig.(\ref{fig:eg3d_glo}). The model $\mathcal{N}$ passes the input conditioning latent $w_i$ to a set of CNN-based synthesis layers \cite{karras2019style} whose output feature maps are used to construct a $\mathrm{k}$-channel tri-plane. The sampled points on each ray are used to extract the tri-plane features and aggregate the $\mathrm{k}$-channel features. Then the tri-plane decoder MLP outputs the scalar density $\sigma$ and color which are alpha-composited by volume rendering to obtain the RGB image. Volume rendering along camera ray $r(t) = O + td$ is:
\vspace{-2mm}
\begin{equation}
    \mathcal{C}_{\text{NeRF}}(r, w) = \int_{b_n}^{b_f} T(t, w) \sigma(r(t), w) \textbf{c}(r(t), \textbf{d}, w) dt \; 
\end{equation}
$$ \text{ where } \; T(t, w) = \text{ exp } \left(-\int_{b_n}^{b_f} \sigma(r(s), w) \right) ds $$
Here, the radiance values can be replaced with the depth $d(x)$ or pixel opacity to obtain the surface depth. During inference, the surface depth map and 2D pixel coordinates are used to obtain the 3D surface points via back-projection. The surface normals can be computed as the first derivative of the density $\sigma$ with respect to the input as follows:
\vspace{-2mm}
$$ \widehat{n}(r, w) = -\int_{b_n}^{b_f} T(t, w) \; \sigma(r(t), w) \; \nabla_{r(t)}(\sigma(r(t), w)) dt $$
\vspace{-1mm}
\begin{equation}
\label{eq:sn}
    n(r, w) = \frac{\widehat{n}(r, w)}{|| \; \widehat{n}(r, w) \; ||_2} 
\end{equation}
Thus from an inference step, an RGB image, surface depth map, 3D surface points and the surface normals of the object instance can be obtained. In Fig.(\ref{fig:eg3dglo_data}), we show the sample reconstruction results for $\mathcal{N}$ on the CelebA-HQ, AFHQv2 and ShapeNet-Cars datasets. In Fig.(\ref{fig:nvs}) we show qualitative results for novel view synthesis with $\mathcal{N}$ trained on SRN-Cars and evaluated on a held-out set of views. Since SRN-Cars is a multi-view dataset, we compare the rendered novel views with their corresponding ground-truth views.

\textbf{Losses.} $\mathcal{N}$ is trained by jointly reconstructing the image and simultaneously optimizing a latent ($w_i$). As noted in \cite{rebain2022lolnerf}, this enables the training loss to be enforced on individual pixels enabling training and inference at arbitrary image resolutions. As depicted in TEGLO Stage-1 in Fig.(\ref{fig:eg3d_glo}), three losses are minimized to train $\mathcal{N}$: $\mathcal{L}_{\text{RGB}}$ is the $\mathcal{L}_1$ reconstruction loss between the pixels from the rendered image and the corresponding pixels from the ground truth image for the object ($o_i$). The $\mathcal{L}_{\text{Perceptual}}$ loss is the LPIPS (Learned Perceptual Image Patch Similarity) loss between rendered image and the ground truth image view. The $\mathcal{L}_{\text{Camera}}$ is the camera prediction $\mathcal{L}_1$ loss between the output of the light-weight camera encoder and the ground-truth camera parameters for the camera pose in order to learn 3D consistent representation of the object $(o_i \in \mathcal{I})$.
\begin{equation}
    \mathcal{L}_{\mathcal{N}} = \mathcal{L}_{\text{RGB}} + \mathcal{L}_{\text{Perceptual}} + \mathcal{L}_{\text{Camera}}
\end{equation}
To train $\mathcal{N}$, we use the single-view image dataset and the approximate pose for each $o_i \in \mathcal{I}$ (Sec.(\ref{sec:experiments})). We train the model for 500K steps using the Adam optimizer \cite{kingma2014adam} on 8 NVIDIA V100 (16 GB) taking 36 hours to complete.

\textbf{Design choices.} As noted in Sec.(\ref{sec:intro}), EG3D \cite{chan2022efficient} shows medium resolution ($512^2$) capacity while using image-space approximations in the super-resolution module which negatively affects the geometric fidelity \cite{skorokhodov2022epigraf}. While EpiGRAF \cite{skorokhodov2022epigraf} uses a patch-based discriminator for pure 3D generation, it is still prone to issues in scaling and training with multi-resolution data. Moreover, adversarial training using discriminators leads to training instability. Different from EG3D and EpiGRAF that use an adversarial training paradigm, $\mathcal{N}$ uses a GLO-based auto-decoder training paradigm which jointly optimizes a latent representation and reconstructs the image enabling arbitrary resolution synthesis - even at large megapixel resolutions - without the constraints of a discriminator. Hence, $\mathcal{N}$ enables 3D representations with geometric fidelity while also benefiting from an efficient tri-plane based representation.

EG3D \cite{chan2022efficient} requires camera pose conditioning for the generator and discriminator to establish multi-view consistency. The limitation of a pose-conditioned generator is that it does not completely disentangle the pose from appearance which leads to artifacts such as degenerate solutions (2D billboards), or expressions such as the eye or smile following the camera. Since $\mathcal{N}$ optimizes a latent representation of an object to reconstruct it, we observe that the generator does not require camera pose conditioning and simply using a light-weight camera predictor network and training with a camera prediction loss ($\mathcal{L}_{\text{Camera}}$) is sufficient to learn 3D consistent representations.

\subsection{TEGLO Stage 2: Dense correspondences}
\label{sec:texture}
\label{sec:eg3dglo_auv}
\vspace{-0.5mm}
\textbf{Formulation.} We render a multi-view dataset ($\mathcal{D}$) using $\mathcal{N}$ trained on single-view image collections for the task of texture representation. We denote each object $e_i \in \mathcal{D}$ comprising of five views: $e_i = \{ v_f, v_l, v_r, v_t, v_b \}$ where $v$ denotes the view, and the sub-scripts ($j \; \text{for all} \; v_j$) denote frontal, left, right, top and bottom poses respectively (refer Fig.(\ref{fig:eg3dglo_data})). In $\mathcal{D}$, each view $v_j \in e_i$ includes the depth map ($\widehat{d}_j$), RGB image ($\widehat{r}_j$), surface normals ($\widehat{s}_j$), 3D surface points ($\widehat{p}_j$), and the optimized latent, $w_i$, which is identical for views of $e_i$ as it is independent of camera pose (Fig.(\ref{fig:eg3dglo_data})). For TEGLO Stage 2, we use  $ \{ \{\widehat{r}_j, \widehat{s}_j, \widehat{p}_j\} \in v_j, w_i \} \in e_i \}$.


Learning dense pixel-level correspondences across multiple views of an object is the task of locating the same 3D coordinate point in a canonical coordinate space. Inspired by surface fields \cite{goli2022nerf2nerf}, we aim to learn dense correspondences using the 3D surface points extracted from $\mathcal{N}$ by back-projecting the depth ($\widehat{d}_j$) and pixel coordinates. Inspired by CoordGAN \cite{mu2022coordgan} and AUVNet \cite{chen2022auv}, we propose a dense correspondence learning network in TEGLO Stage-2 trained in an unsupervised manner learning an aligned canonical coordinate space to locate the same 3D surface point across different views ($v_j$) of the same object ($e_i$).




\textbf{Network architecture.} TEGLO Stage-2 is represented in Fig.(\ref{fig:eg3d_glo}). The architecture consists of a latent mapping network ($\mathcal{L}$), a dense correspondence network ($\mathcal{M}$) and a basis network ($\mathcal{C}$) - all of which are MLP networks. The 3D surface points ($\widehat{p}_j$) from $v_j \in e_i$) are mapped to a 2D canonical coordinate space conditioned on a shape code mapped from the optimized latent $w_i$ for $e_i$. We use a Lipschitz regularization \cite{liu2022learning} for each MLP layer in the dense correspondence network ($\mathcal{M}$). The latent mapping network ($\mathcal{L}$) is a set of MLP layers that takes the $w_i$-latent for $e_i$ as input and predicts a shape-code for conditioning the dense correspondence network $\mathcal{M}$, and coefficients for the deformed basis. Previous work \cite{turk1991eigenfaces, chen2022auv} show that if the input is allowed to be represented as a weighted sum of basis images, \ie to obtain a deformed basis before decomposition, then the 2D canonical coordinate space will be aligned. The basis network ($\mathcal{C}$) is similar to \cite{chen2022auv} and uses the predicted coefficients to decompose the deformed coordinate points. Thus, $\mathcal{M}$ maps the 3D surface points to an aligned 2D canonical coordinate space, enabling the learning of dense correspondences using the $p_j \in \mathcal{S}$ extracted from $\mathcal{N}$. Next, the basis network takes the 2D canonical coordinates as input to predict the deformed basis $\mathcal{B}$. Then, $\mathcal{B}$ is weighted with the predicted coefficients to decompose the basis into the 3D surface points ($p_j$), surface normals ($s_j$) and color ($r_j$).

\textbf{Losses.} TEGLO Stage-2 is trained using three $\mathcal{L}_2$ reconstruction losses: the $\mathcal{L}_{\text{RGB}}$ loss between the rendered RGB image $\widehat{r}_j$ and the predicted RGB image $r_j$; the $\mathcal{L}_{\text{Normals}}$ loss between the rendered surface normals $\widehat{s}_j$ and the predicted surface normals $s_j$; $\mathcal{L}_{\text{Coord}}$ loss between the extracted 3D surface points $\widehat{p}_j$ and the predicted 3D surface points $p_j$. Hence, the total training loss for TEGLO Stage-2 is:
\vspace{-1mm}
\begin{equation}
    \mathcal{L}_{\text{Stage2}} = \mathcal{L}_{\text{RGB}} + \mathcal{L}_{\text{Normals}} + \mathcal{L}_{\text{Coord}}
    \vspace{-1mm}
\end{equation}
To train TEGLO Stage-2, we use the rendered dataset $\mathcal{D}$ consisting of 1000 objects with five views per object and the optimized latent for each identity. The networks are trained using $\mathcal{L}_{\text{Stage2}}$ loss for 1000 epochs using the Adam \cite{kingma2014adam} optimizer to learn dense correspondences across $e_i \in \mathcal{D}$.

\begin{figure*}[t]
  \centering
  \includegraphics[width=0.9\linewidth]{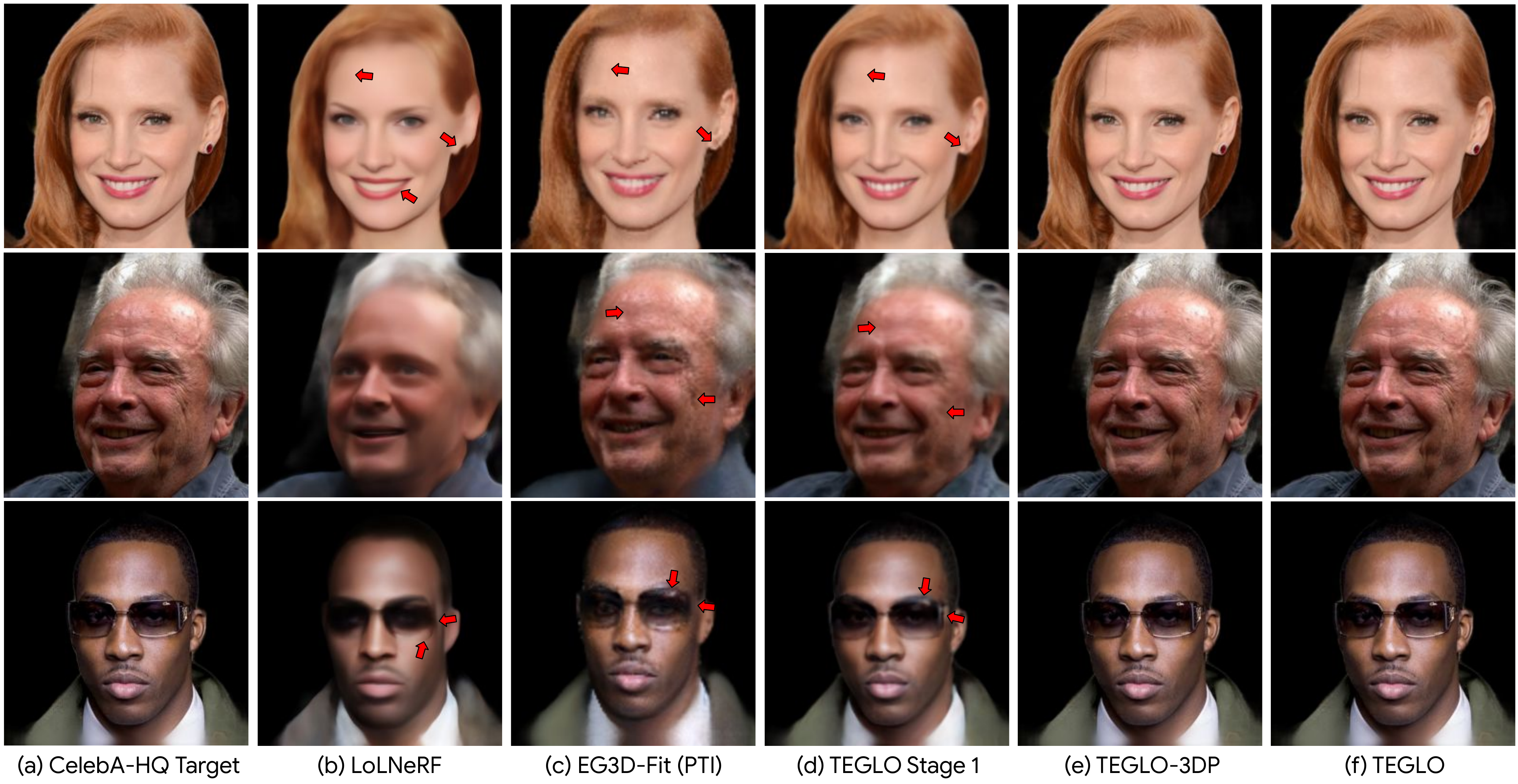}
  \caption{\textbf{Qualitative results} - Comparison with relevant 3D-aware generative baseline methods at $256^2$ resolution for CelebA-HQ.}
  \label{fig:compare}
    \vspace{-3mm}
\end{figure*}

\textbf{Design choices.} We use the optimized $w$-latent from $\mathcal{N}$ for learning the shape code and coefficients for TEGLO Stage-2 because it represents the 3D geometry and appearance information for object ($e_i$) independent of camera pose. We observe that using a Lipschitz regularization for every MLP layer in $\mathcal{M}$ suitably regularizes the network to deform the input surface points $\widehat{s}_j$. Interestingly, our experiments show that simply reconstructing the 3D surface points instead of the color, surface points and surface normals also leads to learning reasonable dense pixel-level correspondences. We show qualitative results for TEGLO Stage-2 trained using only $\mathcal{L}_{\text{Coord}}$ loss in Fig.(\ref{fig:compare}) as TEGLO-3DP.



\begin{figure}[t]
  \centering
  \includegraphics[width=0.99\linewidth]{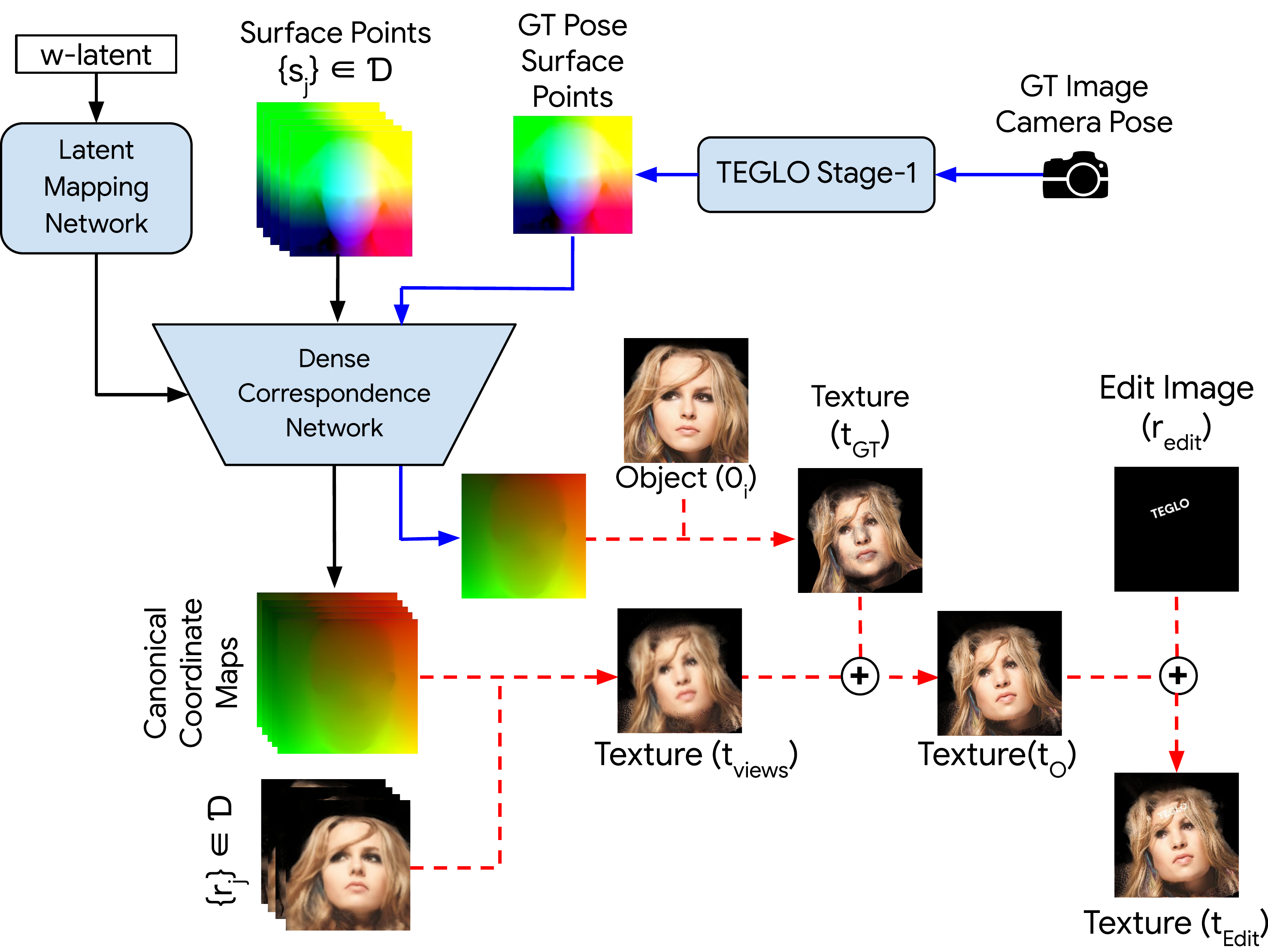}
  \caption{\textbf{Inference} - TEGLO texture extraction for texture transfer and editing. Red arrows indicate the use of a K-d tree to store the texture. Blue arrows indicate the use of input image pixels.}
  \label{fig:texture-infer}
  \vspace{-2mm}
\end{figure}

\begin{figure}[t]
  \centering
  \scalebox{0.99}{
  \includegraphics[width=0.99\linewidth]{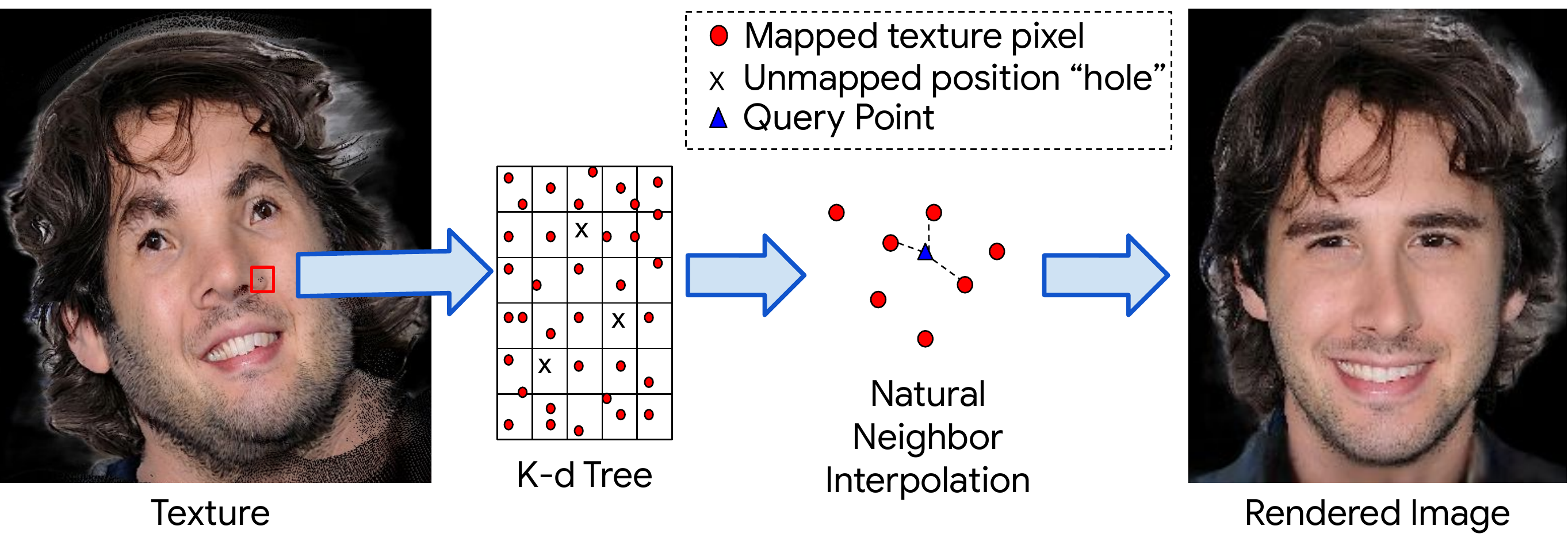}
  }
  \caption{\textbf{Interpolating textures with sparse ``holes"} - Depicting the KD-Tree and Natural Neighbor Interpolation (NNI) to interpolate ``holes" (if any) in the texture for novel view synthesis.}
  \label{fig:kdtree}
  \vspace{-5mm}
\end{figure}

\vspace{-1mm}
\subsection{Inference.}
\label{sec:texture-3d}
\vspace{-1.5mm}
\textbf{Extracting the texture.} After training TEGLO Stage-2, we use the learned dense correspondences to extract a texture map for every object $o_i \in \mathcal{I}$. We use the pose of the target image $o_i$ to extract the 3D surface points from $\mathcal{N}$ and use it to map the image pixels to the 2D canonical coordinate space. We denote this as texture $t_{GT}$. Similarly, we use $\mathcal{M}$ to map the respective RGB values from $\{ v_f, v_l, v_r, v_t, v_b \} \in e_i$ using the corresponding 3D surface points ($s_j$) from all five views to the 2D canonical coordinate space. We denote this as texture $t_{\text{views}}$. Thus, textures $t_{GT}$ and $t_{\text{views}}$ store a mapping \ie the canonical coordinate point and the corresponding RGB values. The procedure is represented in Fig.(\ref{fig:texture-infer}) and textures are depicted in Fig.(\ref{fig:complex}) and Fig.(\ref{fig:editing}). In Fig.(\ref{fig:texture-infer}) $t_{O}$ represents the texture obtained by combining $t_{GT}$ and $t_{\text{views}}$. We store this mapping in a K-d tree which enables us to index into the textures using accurate floating point indices to obtain the RGB values. The K-d tree allows querying with canonical coordinate points to extract multiple neighbors and enables TEGLO to be robust to sparse ``holes" in the texture as depicted in Fig.(\ref{fig:kdtree}).

\textbf{Novel view synthesis.} For rendering novel views of $o_i$, we extract the 3D surface point for the pose from $\mathcal{N}$ and obtain the canonical coordinates from $\mathcal{M}$. For each 2D canonical coordinate point $c_k$, we query the K-d tree for three natural neighbors and obtain indices for the neighbors which are used to obtain the respective RGB values. Natural Neighbor Interpolation (NNI) \cite{sibson1981brief} enables fast and robust reconstruction of a surface based on a Dirichlet tesselation - unique for every set of query points - to provide an unambiguous interpolation result. We simplify the  natural neighbor interpolation (NNI) based only on the distances of the points $c_k$ in the 2D canonical coordinate space to obtain the RGB values from the stored texture. The robust and unambiguous interpolation enables TEGLO to effectively map the ground-truth image pixels from the input dataset $\mathcal{I}$ onto the geometry for novel view synthesis. To extract the Surface Field $\mathcal{S}$, we render $e_i$ from five  camera poses causing potential camera pose biases that may lead to sparse ``holes" in the texture. Our formulation uses the K-d tree and NNI to interpolate and index into textures with sparse ``holes". In Fig.(\ref{fig:kdtree}), each cell in the 5x6 grid represents a discrete pixel in the texture space and the red dot represents a canonical coordinate point. There are three issues that may arise:

\begin{enumerate}[noitemsep,topsep=0pt,itemsep=0pt]
\vspace{-0.75mm}
    \item The canonical coordinate points may not be aligned to the pixel centers and storing them in the discretized texture space may lead to imprecision.
    
    \item There may be multiple canonical coordinates mapped to a discrete integral pixel wherein some coordinates may need to be dropped for an unambiguous texture indexing - leading to loss of information.
    
    \item Some pixels may not be mapped to by any canonical coordinates, creating a ``hole" in discretized space. This is represented by ``\textbf{X}" in the grid in Fig.(\ref{fig:kdtree}). 
\end{enumerate}
\vspace{-0.2mm}
K-d tree allows extracting multiple neighbors by querying with canonical coordinate points and also enables indexing the texture using floating point values. Hence, using a K-d tree to store the texture helps address (1) and (2). Further, using a K-d tree in conjunction with Natural Neighbor Interpolation (NNI) effectively addresses (3). We include more details in the supplementary material. 


\textbf{Texture editing.} Texture editing is represented by $t_{\text{Edit}}$ in Fig.(\ref{fig:texture-infer}). We create the edits on a blank image the same size as that of $t_{\text{O}}$ and denote it as $r_{\text{edit}}$. The edit image $r_{\text{edit}}$ is taken to be in the canonical coordinate space and hence directly indexed into the K-d tree to be overlay on $t_{O}$. Note that the we do not apply any constraint on the texture space represented and hence the texture may be visually aligned to a non-frontal canonical pose as is the case in Fig.(\ref{fig:complex}) and Fig.(\ref{fig:editing}). The final texture with the edit $t_{\text{Edit}}$ is created by combining $t_{O}$ and $r_{\text{edit}}$. Qualitative results for texture edits are depicted in Fig.(\ref{fig:teaser}) and Fig.(\ref{fig:editing}).



%% file: Sections/4_Experiments.tex
\textbf{Datasets.} We train TEGLO with single-view image datasets such as FFHQ \cite{karras2019style}, CelebA-HQ \cite{liu2015faceattributes, karras2017progressive} and AFHQv2-Cats \cite{karras2021alias, choi2020stargan}. To obtain the approximate camera pose, we follow \cite{rebain2022lolnerf} by first using an off-the-shelf face landmark predictor MediaPipe Face Mesh \cite{MediaPipe} to extract landmarks appearing at consistent locations. Then, we use a shape-matching least-squares optimization to align the landmarks with 3D canonical landmarks to obtain the approximate pose. We also use a multi-view image dataset - ShapeNet-Cars \cite{shapenet2015, chang2015shapenet} with results in Fig.(\ref{fig:teaser}) and Table.(\ref{tab:teglo_test_2}).

\begin{figure}[t]
  \centering
  \includegraphics[width=0.99\linewidth]{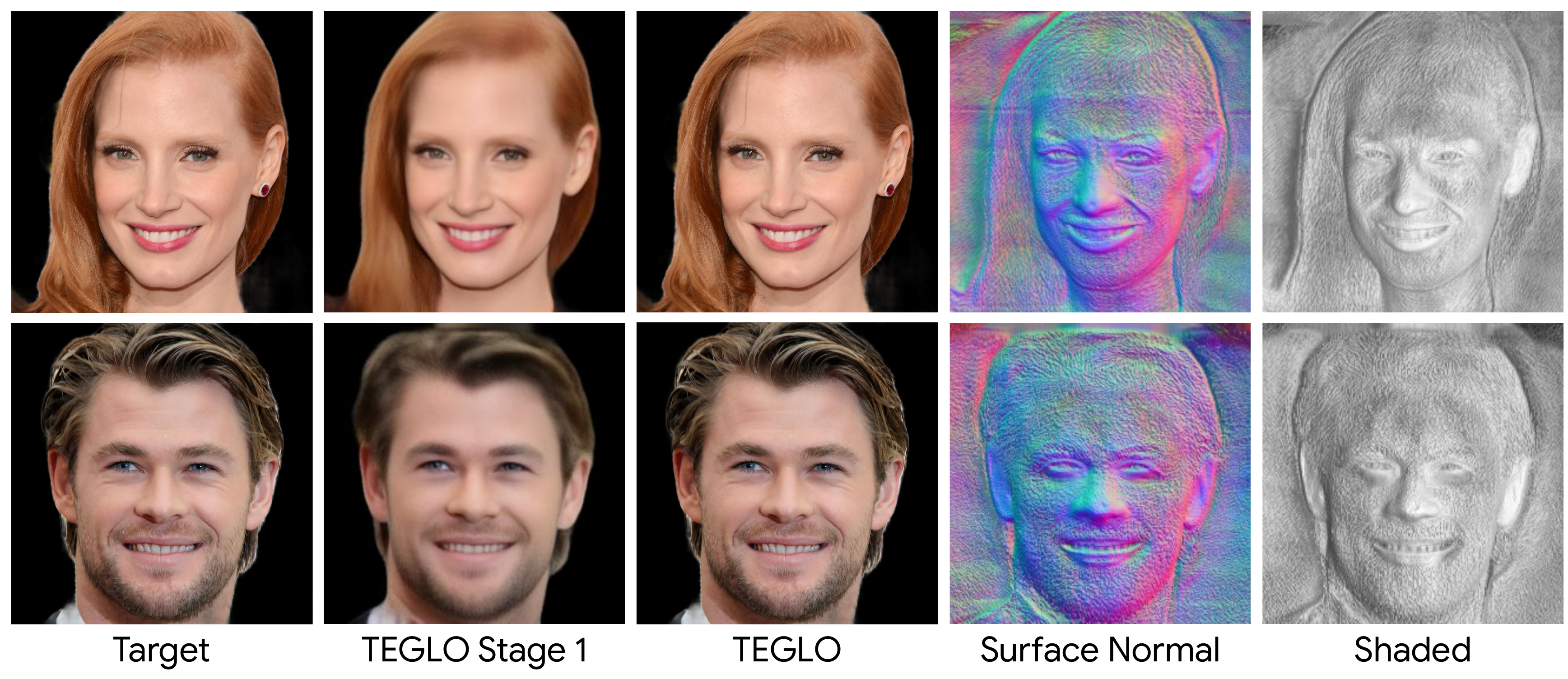}
  \caption{\textbf{Single view 3D reconstruction} - Results for TEGLO trained on FFHQ data and evaluated on CelebA-HQ image targets.}
  \label{fig:single-view-3d}
  \vspace{-3mm}
\end{figure}

\begin{figure}[t]
  \centering
  \includegraphics[width=0.99\linewidth]{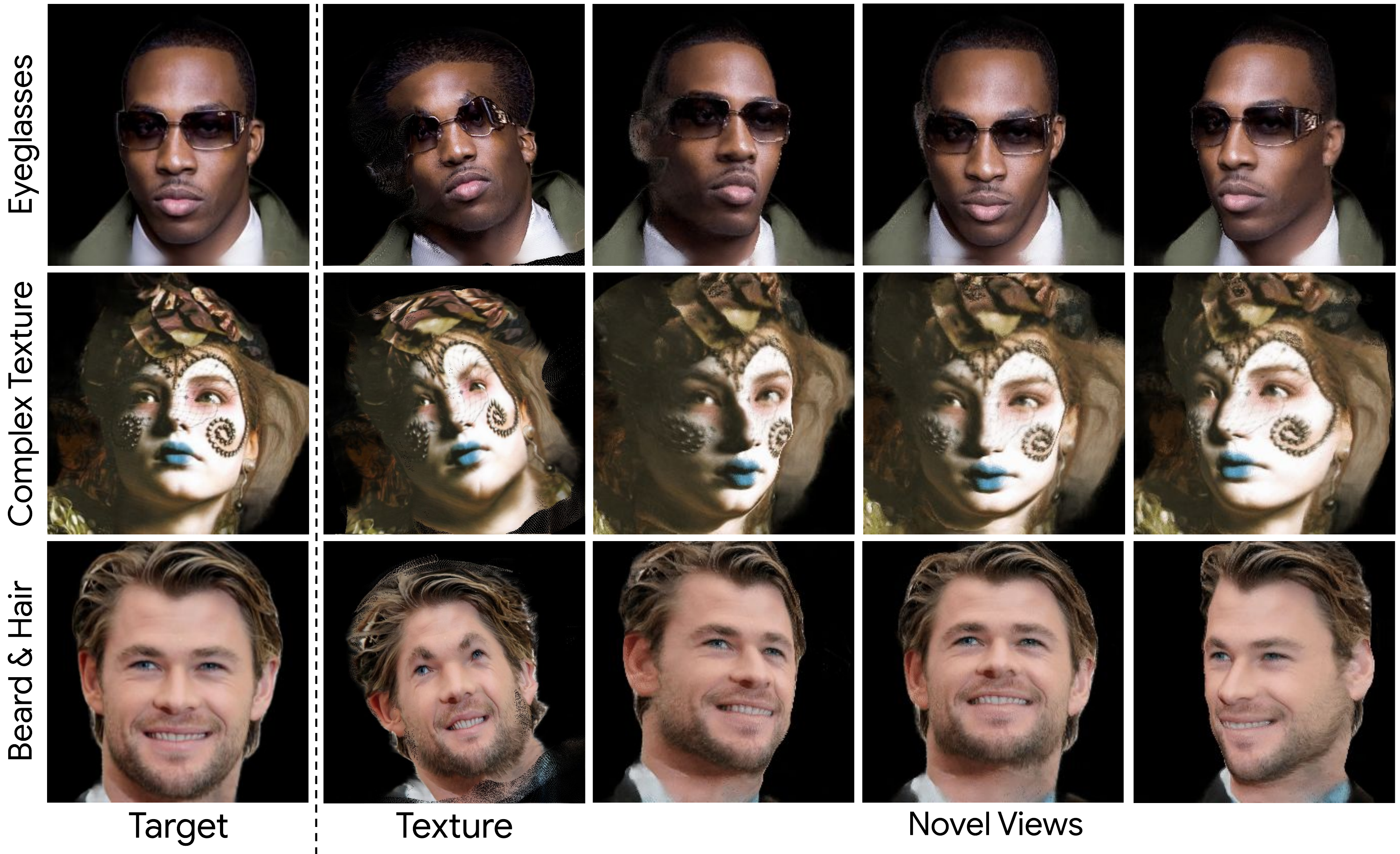}
  \caption{\textbf{Results for complex texture and geometry} - Qualitative results for texture representation and novel view synthesis with complex image samples. Compare row-1, row-2 with Fig.(24) in \cite{cao2022authentic} and row-3 with Fig.(25) in \cite{li2020dynamic})}
  \label{fig:complex}
  \vspace{-4mm}
\end{figure}

\textbf{3D reconstruction.} We evaluate TEGLO on the task of reconstructing the input image in the same pose and show a comparison with baseline methods. We report quantitative results for train data reconstruction in Table.(\ref{tab:teglo_train}) measuring the PSNR (Peak Signal to Noise Ratio) and LPIPS (Learned Perceptual Image Patch Similarity) metrics for CelebA-HQ and FFHQ. We observe similar results for LoLNeRF and TEGLO Stage-1 at $128^2$ resolution. However, as expected, TEGLO attains $89.5$ dB PSNR and $7.4e$-7 for LPIPS. We report quantitative results for test data reconstruction at $256^2$ resolution from a held-out set from the dataset for CelebA-HQ and FFHQ in Table.(\ref{tab:teglo_test}), and AFHQv2-Cats and SRN-Cars in Table.(\ref{tab:teglo_test_2}). We observe that TEGLO attains near-perfect reconstruction of test data attaining $\geq 67.5$ dB PSNR and $\leq 6.9e$-5 for LPIPS across the datasets. 



We depict qualitative results for CelebA-HQ in Fig.(\ref{fig:compare}) where the red arrows indicate missing high frequency details for 3D reconstruction. For EG3D-Fit, we invert the image into the EG3D \cite{chan2022efficient} latent space and perform Pivotal Tuning Inversion (PTI) \cite{roich2022pivotal} for the single-view images. We observe missing details in LoLNeRF \cite{rebain2022lolnerf}, EG3D-Fit \cite{chan2022efficient} and TEGLO stage 1 specifically related to details such as jewelry, skin wrinkles, eyeglass translucency, eyeglass frames, hair strands etc. As expected, the results for TEGLO and TEGLO-3DP (where TEGLO Stage-2 is trained with only surface point supervision) include the high frequency details missed by baseline methods demonstrating near perfect reconstruction. In Fig.(\ref{fig:complex}), we show qualitative results including the texture image ($t_O$) for complex appearance and geometry such as multi-view consistent eyeglasses, 3D make-up and hair. Compared with Fig.(24) in \cite{cao2022authentic}, we show improved multi-view consistent results for eyeglasses in row-1, and 3D make-up in row-2. Compared with Fig.(25) in \cite{li2020dynamic}, we show multi-view consistent representations for beard that the baseline method \cite{li2020dynamic} was unable to model.



\begin{table}[t]
\centering
    \caption{\textbf{Reconstruction of train images} - Quantitative comparison on training data reconstruction at $128^2$ resolution.}
    \label{tab:teglo_train}
    \scalebox{0.9}{
    \begin{tabular}{lccc}
        \hline
        Method & PSNR ($\uparrow$) & LPIPS ($\downarrow$)  \\ \hline
        $\pi$-GAN \cite{chan2021pi} (CelebA) & 23.5 & 0.226 \\
        LoLNeRF \cite{rebain2022lolnerf} (FFHQ) & 29.0 & 0.199 \\
        LoLNeRF \cite{rebain2022lolnerf} (CelebA-HQ) & 29.1 & 0.197 \\
        ABC \cite{rebain2022attention} (CelebA-HQ) & 26.3 & - \\
        TEGLO Stage 1 (FFHQ) & 29.0 & 0.294 \\
        TEGLO Stage 1 (CelebA-HQ) & 28.9 & 0.317 \\
        TEGLO (CelebA-HQ) & \textbf{89.5} & \textbf{2.3e-7} \\
        \hline
    \end{tabular}
    }
    \vspace{-2mm}
\end{table}


\begin{table}[t]
\centering
    \caption{\textbf{Reconstruction of test images} - Quantitative comparison on test data reconstruction.} \vspace{1mm}
    \label{tab:teglo_test}
    \setlength{\tabcolsep}{5pt}
    \scalebox{0.875}{
    \begin{tabular}{lccc}
        \hline
        Method & Res. & PSNR ($\uparrow$) & LPIPS ($\downarrow$)\\ \hline
        $\pi$-GAN \cite{chan2021pi} (CelebA) & $256^2$ & 21.8 & 0.412 \\
        LoLNeRF \cite{rebain2022lolnerf} (FFHQ) & $512^2$ & 25.3 & 0.491\\
        LoLNeRF \cite{rebain2022lolnerf} (CelebA-HQ) & $256^2$ & 26.2 & 0.363\\
        TEGLO Stage 1 (FFHQ) & $256^2$ & 27.3 & 0.334 \\
        TEGLO Stage 1 (CelebA-HQ) & $256^2$ & 27.5 & 0.260 \\
        \hline
        TEGLO (FFHQ)& $256^2$ & \textbf{84.9} & \textbf{2.1e-6} \\
        TEGLO (CelebA-HQ)& $256^2$ & \textbf{86.2} & \textbf{7.4e-7} \\
        \hline
        TEGLO (CelebA-HQ) & $512^2$ & 82.6 & 4.4e-6 \\
        TEGLO (CelebA-HQ) & $1024^2$ & 74.7 & 6.9e-5 \\

        \hline
    \end{tabular}
    }
    \vspace{-2mm}
\end{table}

\begin{table}[t]
\centering
    \caption{\textbf{Comparing with GLO baselines} - Quantitative results for test set reconstruction in PSNR at $256^2$ resolution with Generative Latent Optimization baselines. (We report the LoLNeRF result on SRN-Cars from the "Concatenation" baseline in \cite{rebain2022attention}).}
    \label{tab:teglo_test_2}
    \scalebox{0.875}{
    \begin{tabular}{cccc}
        \hline
        \multirow{1}{*}{Method} & \multicolumn{1}{c}{AFHQv2-Cats \cite{karras2021alias}} & \multicolumn{1}{c}{SRN-Cars \cite{chang2015shapenet}}
        \\ \hline
        LoLNeRF \cite{rebain2022lolnerf} & 24.94 & 25.80\\
        ABC \cite{rebain2022attention} & - & 29.10\\
        TEGLO Stage-1 & 29.26 & 30.48\\
        TEGLO & \textbf{87.38} & \textbf{67.52}\\
        \hline
    \end{tabular}
    }
    \vspace{-4mm}
\end{table}

\textbf{Single-view 3D reconstruction.} It is the task of representing an in-the-wild or out-of-distribution image using a trained network. The qualitative results for TEGLO trained on FFHQ \cite{karras2019style} data for single-view 3D reconstruction on samples from CelebA-HQ are in Fig.(\ref{fig:single-view-3d}). Previous work such as AUVNet \cite{chen2022auv} require additional training of a ResNet-18 \cite{he2016deep} for the image encoder and IM-Net \cite{chen2019learning} for the shape decoder followed by ray marching to obtain the mesh to represent the image while methods such as EG3D \cite{chan2022efficient} require PTI (Pivotal Tuning Inversion \cite{roich2022pivotal}) fine-tuning to represent the image. For single-view textured 3D representation in TEGLO, we simply invert the image into the latent and do not require any fine-tuning.

Reconstructing single-view images at arbitrary resolutions while preserving 3D consistency is very desirable for many applications. However, EG3D \cite{chan2022efficient} has a limitation in performing this task because its generator is conditioned on the camera intrinsic and extrinsic parameters, leading to a ``baked-in" training image resolution. As TEGLO does not condition on the camera, it enables single-view 3D reconstruction and novel view synthesis at arbitrary resolutions without requiring re-training for different resolutions.


\begin{figure}[t]
  \centering
  \includegraphics[width=0.99\linewidth]{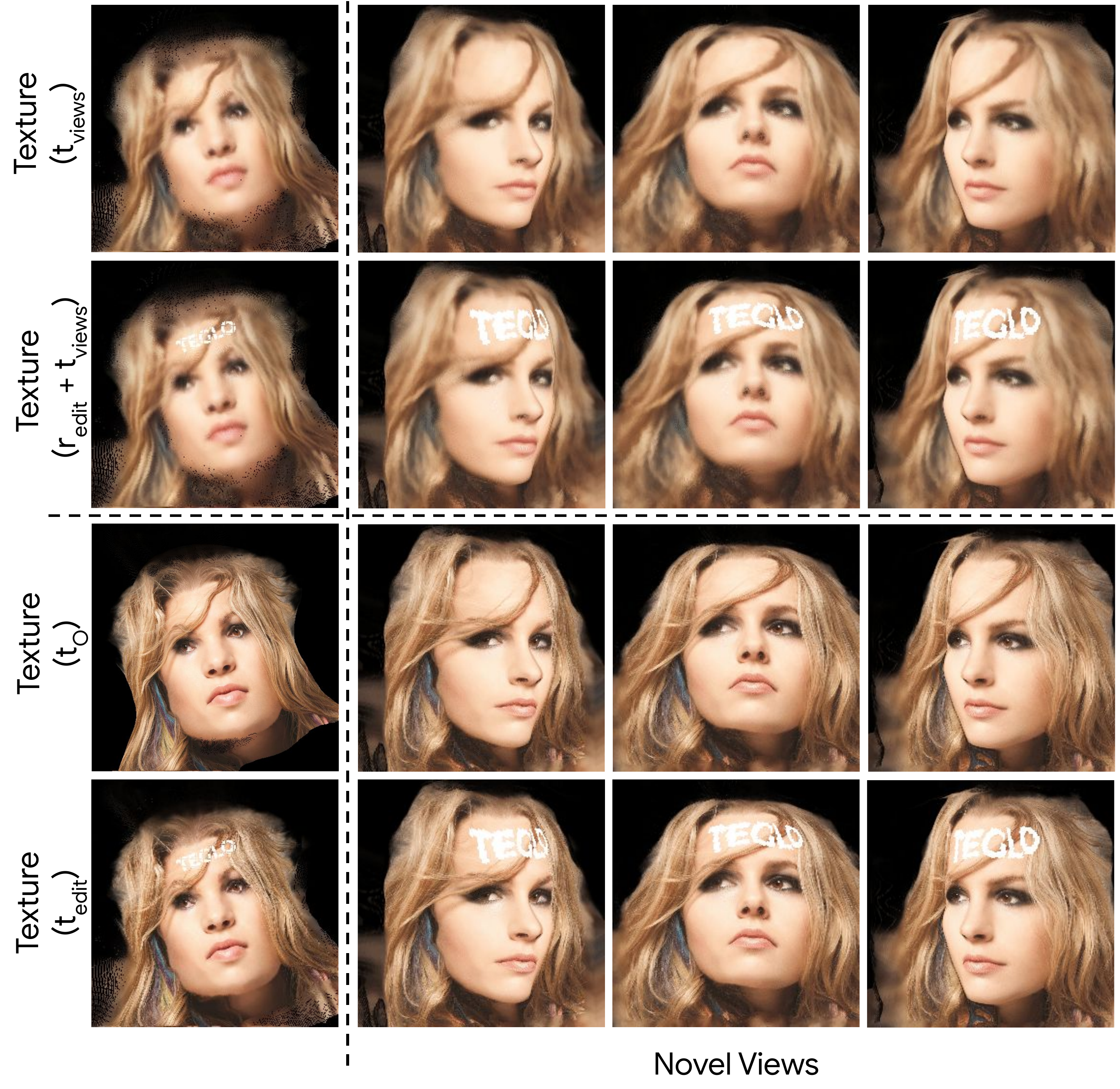}
  \caption{\textbf{Texture editing} - Qualitative results for texture edits.}
  \label{fig:editing}
  \vspace{-4mm}
\end{figure}

\textbf{Texture editing.} In Sec.(\ref{sec:texture-3d}), we describe the procedure to edit textures using TEGLO. The qualitative results with texture editing for CelebA-HQ \cite{liu2015faceattributes, karras2017progressive} are depicted in Fig.(\ref{fig:editing}) and for AFHQv2-Cats and ShapeNet-Cars in Fig.(\ref{fig:teaser}). Our edits are class-specific and target image agnostic because edits are performed in the canonical space. Previous work, NeuMesh \cite{yang2022neumesh} requires spatial-aware fine-tuning and mesh guided texture editing for precise transfer. However, TEGLO simply maps a texture edit image of the same size as the texture into the K-d tree with an overlay of the pixels in the earlier texture (\ie obtaining $t_{\text{Edit}}$) - precisely transferring the edit without requiring any optimization strategies. Further results are in the supplementary.


\textbf{Texture transfer.} As discussed in Sec.(\ref{sec:texture-3d}), the extracted textures are aligned in a canonical coordinate space and allows transferring textures across different geometries. We demonstrate texture transfer across different geometries in Fig.(\ref{fig:texture-transfer}). Here, row-1 represents the target image from CelebA-HQ for the geometry learned by TEGLO Stage-1 and column-1 represents the textures (stored in a K-d tree) extracted after TEGLO Stage-2. We observe realistic texture transfer despite arbitrary camera biases in rendering $\mathcal{D}$ mitigated by using the K-d tree and NNI. (Fig.(\ref{fig:kdtree})).

%% file: Sections/5_Discussion.tex
While TEGLO enables near perfect 3D reconstruction of objects from single-view image collections, it requires non-trivial computational resources to train TEGLO Stage-1, render a dataset with five views of the object, train TEGLO Stage-2 and then use the input image surface points to extract the texture. We hope that future work can simplify the framework with an elegant end-to-end formulation. 

A trivial next step would be to use StyleGANv2 \cite{karras2019style} to generate high quality textures for texture transfer and editing. TEGLO could enable 3D full-body avatars from single views with an unprecedented amount of detail preservation extending methods such as PIFu \cite{saito2019pifu}. Future work could explore representing light stage data via NeRFs with high frequency details across different camera capture angles in an illumination invariant manner using 3D surface points. 





\begin{figure}[t]
  \centering
  \includegraphics[width=0.92\linewidth]{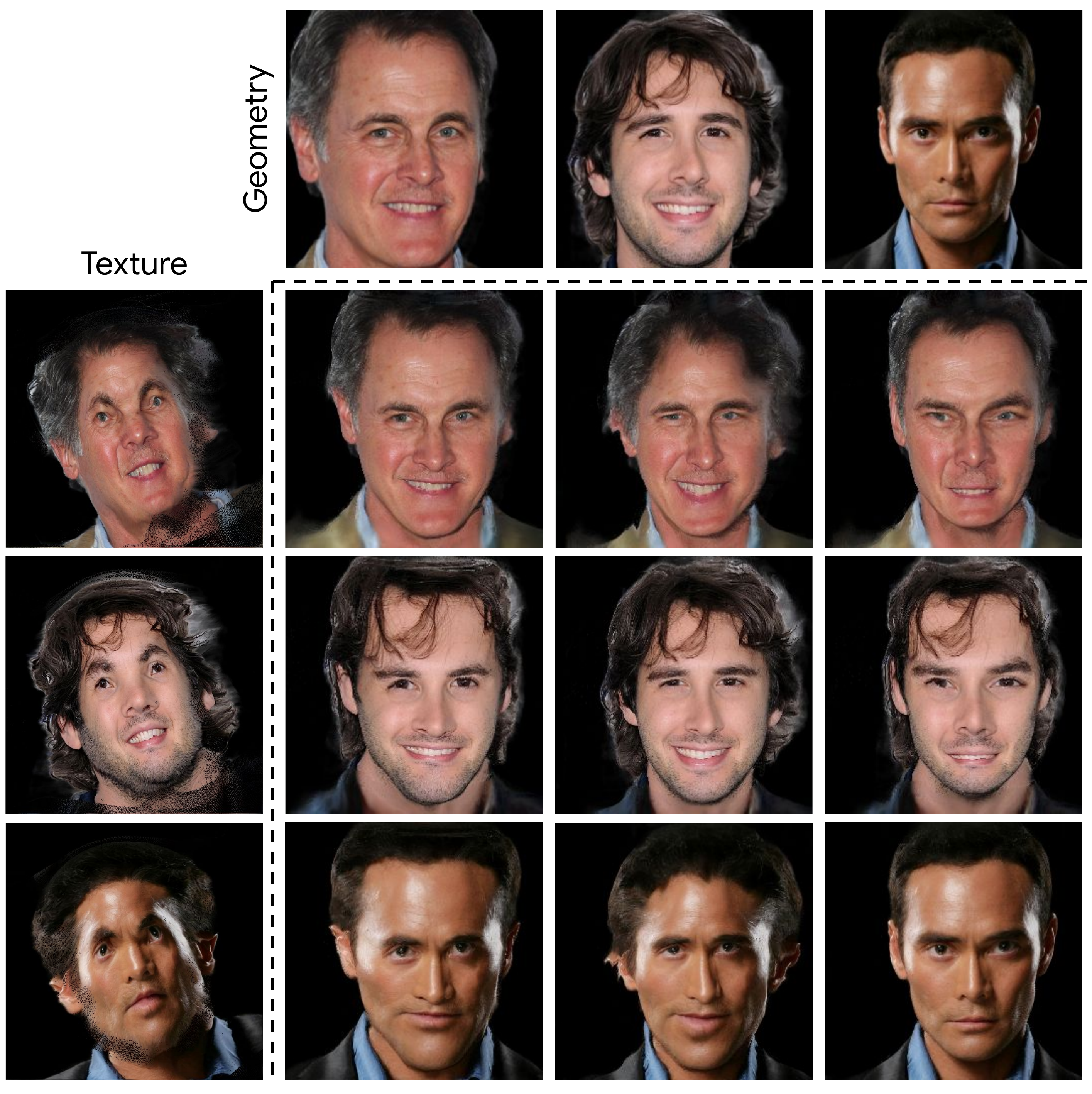}
  \caption{\textbf{Texture transfer} - Qualitative results for texture transfer with CelebA-HQ. (Top row shows CelebA-HQ image targets).}
  \label{fig:texture-transfer}
  \vspace{-3mm}
\end{figure}

%% file: Sections/6_Conclusion.tex
In this work, we present TEGLO for high-fidelity canonical texture mapping from single-view images enabling textured 3D representations from class-specific single-view image collections. TEGLO consists of a conditional a NeRF and a dense correspondence learning network that enable texture editing and texture transfer. We show that by effectively mapping the input single-view image pixels onto the texture, we can achieve near perfect reconstruction ($\geq 74$ dB PSNR at $1024^2$ resolution). TEGLO allows single-view 3D reconstruction by inverting the single-view image into the latent table without requiring any PTI or fine-tuning.